\documentclass[final,5p,times,twocolumn]{elsarticle}

\usepackage{hyperref}
\usepackage{graphicx}

\usepackage{tabulary}
\usepackage{booktabs}
\usepackage{amssymb}

\journal{Knowledge Based Systems}

\begin{document}

\begin{frontmatter}

\title{A Multi-Disciplinary Review of Knowledge Acquisition Methods: From Human to Autonomous Eliciting Agents}

\author{George Leu \fnref{fn1}} %and Hussein Abbass}
%\author{Author One\corref{cor1}\fnref{fn1}}
%\ead{email@uni.edu}
%\cortext[cor1]{Corresponding author. Email: G.Leu@adfa.edu.au Ph: +61262688424}
\fntext[fn1]{Corresponding author: E-mail: G.Leu@adfa.edu.au Ph: +61262688424}

\author{Hussein Abbass}

\address{School of Engineering and IT, UNSW Canberra, Australia}

\begin{abstract}

This paper offers a multi-disciplinary review of knowledge acquisition methods in human activity systems. The review captures the degree of involvement of various types of agencies in the knowledge acquisition process, and proposes a classification with three categories of methods: the human agent, the human-inspired agent, and the autonomous machine agent methods. In the first two categories, the acquisition of knowledge is seen as a cognitive task analysis exercise, while in the third category knowledge acquisition is treated as an autonomous knowledge-discovery endeavour. The motivation for this classification stems from the continuous change over time of the structure, meaning and purpose of human activity systems, which are seen as the factor that fuelled researchers' and practitioners' efforts in knowledge acquisition for more than a century.

We show through this review that the KA field is increasingly active due to the higher and higher pace of change in human activity, and conclude by discussing the emergence of a fourth category of knowledge acquisition methods, which are based on red-teaming and co-evolution.
\end{abstract}


\end{frontmatter}

\section*{List of abbreviations}

\begin{tabular}{l l}
	KA & Knowledge Acquisition \\
	HAS & Human Activity System \\
	CTA & Cognitive Task Analysis \\
	SME & Subject Matter Expert \\
	KD & Knowledge Discovery \\
	EC & Evolutionary Computation \\
	CRT & Computational Red Teaming \\

\end{tabular}

\section{Introduction}

Knowledge Acquisition (KA) refers in a very broad view to gaining understanding about the processes underlying the observable behaviour of an entity. The immediate output of KA, the knowledge, is a representation of the real phenomenon at the level of detail and abstraction required by the purpose of the KA exercise. The representation takes the form of an ontological construct, i.e. a set of concepts considered necessary and sufficient to capture the understanding about the real phenomenon, which offers the possibility to re-instantiate it (replicate it in a different context), to improve it, or to further communicate the understanding about it to peers. The range of purposes for knowledge acquisition exercises is very broad, from the basis of learning in itself, to the creation of computational models and applications that improve the behaviours of the entities under investigation or solve problems on behalf of them (e.g. knowledge-based systems, expert systems).

The entities the KA can be performed on fall into three major categories - natural, man-made and humans - further referred to throughout the paper as natural systems, technical systems, and human activity systems. An example of KA applied to natural systems can be the weather cycle in a certain region of the Earth, which one needs to understand in order to ensure safe aircraft operation over that region in different periods of the year. The tools available for gaining understanding about the weather are observations and measurements. The weather cycle is observed, measurements are taken on some relevant aspects such as air speed, temperature, pressure or humidity, and records of these observations and measurements are analysed and structured in order to understand how and why the weather behaves the way it does. This further allows the representation of the weather cycle in a manner that can be communicated to and used by aircraft operators. The same tools, i.e. observations and measurements, are available when applying KA to man-made technical systems, like in the case of an aircraft life-cycle. During its life-cycle, from design to manufacturing, operation and decommissioning, an aircraft is under permanent observation, and large amounts of measurements are performed in order to gain understanding on all possible aspects that allow normal operation. For example, before commissioning into operation, the design is tested in simulations, then in controlled realistic environments (e.g. wind tunnels) and finally in real flight tests in order to gain understanding about how all system components interact internally within the aircraft and externally with other systems such as the weather or the operators. The resultant knowledge can be used to improve the design if the initial design assumptions are not met, or to release the aircraft into operation and communicate this knowledge to its operators.

Unlike natural and technical systems, in the case of human activity the processes underlying the observable behaviour can be unveiled not only through observations and measurements, but also by asking. Weather can be observed and measured, but cannot be asked about why it is the way it is. Similarly, the operation of an aircraft can be observed and measured, but the aircraft cannot be asked why it manifested a certain behaviour in some particular weather conditions. Humans instead, can be asked and thus knowledge can be elicited through various techniques that are unavailable in the case of natural and technical systems. For example, we assume an activity such as piloting an aircraft. This is a human activity system through that it involves the existence and the interaction of all types of entities: natural - the weather, technical - the aircraft and human - pilot's actions and decision making, with the human being the pertinent entity that steers the whole resultant system and is accountable for its behaviour. The pilot in this case makes use of its knowledge about the natural system, its knowledge about the technical system, and its innate or acquired cognitive-motor and decision-making skills in order to perform the task of flying the aircraft in good conditions. If one intends to improve the skills of this pilot (for purposes such as safety, flight duration, passenger comfort) or to transfer the existing skills to other pilots (through creation of training programs), then it is paramount to gain understanding about how and why the pilot takes decisions and performs various actions, and how are these yielding from the subsequent interactions with the natural and technical counterparts (the weather and the aircraft). Further, it is paramount to create and commit to an ontological construct that represents this understanding effectively, in order to be able to use it for fulfilling the established purposes.

In the light of the above examples, in this paper we concentrate on knowledge acquisition in relation to the generalised concept of human activity (described in the third example - the pilot), a research direction in which knowledge acquisition is employed as a facilitator for finding ways to improve human performance in various tasks in real-world contexts \citep{Militello2008,Roth2014}. Historically, this research direction emerged in response to the need for improving the ``workplace'', where the word workplace has a broad meaning, referring to the physical work-place itself, but also encompassing the tasks performed by humans as part of their lucrative activity, their proficiency in accomplishing those tasks, their interaction with the technology they use in support of that lucrative activity, and the artefacts resulting from their activity. Roth et al. \citep{Roth2014} see knowledge acquisition through a cognitive task analysis (CTA) lens, and note that KA is nowadays an indispensable tool used to understand the ``cognitive and collaborative demands'' that contribute to performance and facilitate the formation of expertise. They also note that KA is used as a support for designing ways to improve individual performance through various forms of training, user interfaces, human-machine interaction or decision-making support systems.

Today KA in relation to human activity systems is addressed in multiple fields of activity. In \textit{Cognitive Systems Engineering} the KA methods are used to analyse the work environment in order to inform the design of various systems, focusing on the integration of humans, technologies and physical work space \citep{Militello2008}. In \textit{Cognitive Work Analysis} \citep{Roth2008,Houghton2015} the KA exercise is used for making real-world constraints more visible to human operators in order for them to make better-informed decisions in unanticipated circumstances. In \textit{Naturalistic Decision Making} \citep{Orasanu2001}, some researchers proposed KA methods for investigating how human decision-making emerges in real world tasks, as a result of time pressure and risk \citep{Schraagen2008,Militello2008}. \textit{Human-Centered Computing} is another major field of research where KA techniques were used as a support for designing technology that amplifies or extend human capabilities \citep{Sharples2002}. KA was also considered essential for general \textit{Knowledge Engineering} \citep{Cooke1994,Crandall2006}, where KA can be used in any or all of the knowledge elicitation, analysis and representation stages. In addition, in \textit{Knowledge Discovery and Data Mining} \citep{Jagielska1999} computational intelligence instantiations of KA exercises are used for autonomous knowledge discovery in problem domains that only offer access to inexact and imprecise artefactual data resulting from human activity systems. While this list is not exhaustive, it shows the magnitude of the KA paradigm and its importance in the investigation of what we can broadly consider, virtually any human activity system.

\begin{figure}[h]
\begin{center}
    \includegraphics[width=0.99\linewidth]{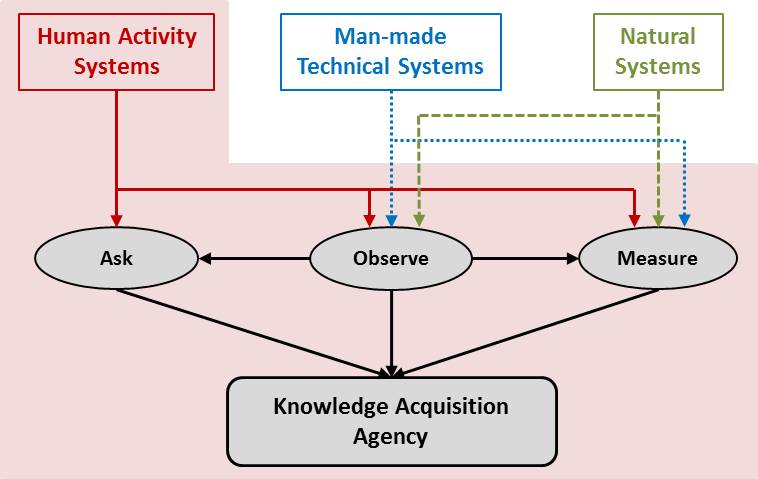}
    \caption{The shaded area shows the position and scope of the review within the larger KA field.}
    \label{Fig:KAfield}
\end{center}
\end{figure}

Figure~\ref{Fig:KAfield} summarises the above discussion in a visual manner, presenting broadly the position and scope of this study within the larger field of knowledge acquisition. More specifically, the review sees the human activity KA literature from an agent perspective and classifies it into three major categories of methods: human agents, human-inspired agents, and machine agents. In the most general view, KA exercises assume various degrees of interaction between one or more elicitor entities and one or more subject matter expert (SME) entities. In the proposed classification the human agent category refers to the classic Cognitive Task Analysis exercises, in which both the eliciting entities and the subject matter expert entities are humans. However, advances in computational intelligence and digital technologies made possible the replacement of either or both sides involved in the CTA exercise with their computer-based counterparts. Thus, the second category, human-inspired agents, consists of the classic CTA exercises in which the human elicitor entities are replaced by computer programs that emulate well-established human elicitors' actions (e.g. automated interviews, automated protocol analysis). The third category, the machine agents, include those KA exercises in which both the eliciting entity and the SME entity are replaced by computer programs with independent behaviour and no domain knowledge regarding the two entities. That means, the eliciting programs do not implement known behaviour of human elicitors, instead, they act autonomously for extracting knowledge from the SME entity. In the same time the SME entities are not any more human experts or known recorded behaviour of them, but rather unstructured and unsorted artefactual data resulted from unattended records of the human activity. Thus, in this category, the KA exercise becomes a Knowledge Discovery (KD) endeavour, in which the machine agents attempt to establish causal relations and meaning in an unstructured collection of artefactual data believed to be related to a task or human activity system of interest.

The paper is organised as follows. In the second section, we present a historical view on KA, which provides an intrinsic chronology-based categorisation of the methods and provides the motivation for this study. Section 3 presents a discussion that highlights the fundamental stages of the KA process and sketches an architectural schema capturing general KA exercises. The fourth section presents the existing classifications of the KA methods and provides the foundation for the following sections. Sections 5, 6 and 7 review the methods according to the three proposed categories, human agents, human-inspired agents and machine agents, respectively. In Section 8 we discuss major findings, the contribution of this paper, and future research directions. Section 9 briefly concludes the study.
 
\section{A historical view}

The KA concept has its roots in a long history of research on human activity, researchers tracing the work in the field to as early as 19th century. A brief historical view can be found in \citep{Militello2008}, while a comprehensive and in-depth discussion can be found in \citep{Hoffman2008} in relation to four major periods acknowledged in the studies related to HAS: pre-1900, 1900-1920's, 1920's - 1960's, and 1970's to present. The first three are related to precursors of KA, while the fourth one represents the modern KA. The pre-1900 period was focused on self-reporting methods based on introspection and retrospection, in which knowledge about task and task performance was obtained from human subjects by asking them to perform the task and then provide retrospective reports. The second period, 1900 to 1920's, was dominated by the concept of ``psychotechnics'' which emerged as a result of the increased pace of industrialisation, which in turn generated the need for improving the human performance at work in order to increase productivity. Studies were concentrated on analysis of tasks, usually physical in nature, and targeted timing and motion during one task, or between several tasks, throughout the work place. Seminal for this period was the time and motion study of Taylor, first published in 1911 \citep{Taylor1967}, where the focus was on understanding how proficient workers performed their tasks. The third period, 1920's to 1960's, was under the influence of large scale manufacturing, when mass production in various industries introduced automation and high productivity assembly lines, changing dramatically the HAS. Task analysis shifted towards the study of human factors, work safety and human-machine systems, where the analysis of human activity was made with the purpose of adapting the tools and processes to the human operators for increasing productivity.

The fourth period, starting in the 1970's, is when KA in human activity systems emerged as a standalone high-level methodology covering all the fields previously addressed by precursor methods \citep{Gallagher1979,Cooke1994}. The technological advances shifted once more the human activities from predominantly physical to predominantly cognitive tasks, and consequently the human-technology interaction and integration problem reached levels unseen before \citep{Militello2008}. We consider that the fourth period marks the most important contributions to the KA body of research, establishing the concept solidly in the scientific world through qualitative and quantitative studies. The fourth period, which continues today, was when most of the human agent (i.e. human-based or manual) KA methods and techniques were proposed and largely accepted by both researchers and practitioners. However, we note that since the early 2000's the amount of new methods and methodologies in the human agent KA area slightly declined, as evident through our inspection of the literature \citep{Cooke1994,Wei2004,Militello2008}, while the amount of work on applications of the existing methods increased significantly. This is, we argue, due to the fact there was yet another shift in the HAS. It was the unprecedented development of the internet, the mobile communications, and lately virtual environments and social media/networks that gave birth to the information age, and transformed human activity systems to such extent that classical human agent KA exercises were no longer efficient or convenient. First, the physical interaction of human elicitors with human SMEs became less and less plausible and possible due to the nature of the new types of human activity. Then, the new range of activities generate behavioural artefacts in quantities and shapes no longer treatable with the capabilities provided by human-based methods.

As a response to the emergence of these new types of human activity, the human-inspired KA agents appeared as one direction of research, trying to implement some of the human agent methods in computer programs in order to replace human elicitors with their automated versions. However, these methods are still computer implementations of classical methods. A second direction of response consisted of methods that diverged from the human-related KA, bringing into the KA field human-independent approaches, the machine agents, which managed to better cope with acquisition of knowledge in cases where the human element became inefficient. This direction is marked by the full inclusion of agent-enabling technologies based on computational intelligence, such as statistical analysis, machine learning or evolutionary computation, into the KA exercises. We consider these two directions as belonging to a fifth significant period in the historical view on KA, starting from the 1990's to present days, which adds to the four historical periods acknowledged in the literature. We also note that despite the reduction in the amount of new methods belonging to the human agent methods, the fourth period did not cease to exist. Research on the classic human agent KA continued until the present days with valuable reiterations of the established methods in new contexts and application fields. Thus, the fifth period, which brings the human-inspired and machine agents to KA comes not as a replacement, but as an addition to existing paradigms, fuelled by the continuous expansion of the concept of human activity.

In the light of the above historical view on the KA research, and motivated by the historical evolution of the concept of human activity, we consider pertinent and useful to discuss the literature based on the three categories we mentioned in the introductory section: the human agents, the human-inspired agents and the machine agents. In order to keep the scientific discourse pertinent and related to the KA concept, we provide in the next section a discussion on the core process of a KA exercise. This discussion will guide our review, and will ensure that the methods we describe throughout the paper are only included if they belong to the core KA process and suit the purpose of the KA exercises in relation with the concept of human activity.

\section{Agent-enabled KA exercises: core process and roadmap}

To date, the literature focused more on the internal mechanism of the KA methods, and less on establishing an accurate description of the KA process, or a relation between the process and the purpose of the exercise \citep{Cooke1994,Wei2004,Crandall2006,Yates2007,Yates2011}. Thus, the choice on how to conduct a KA exercise is strongly dependant on the purpose of the exercise. In \citep{Roth2014} the authors consider, from a CTA perspective, that KA is still a task-related paradigm, ``with no best practice, or general methodology in place''. Also, from a KD perspective Kurgan and Musilek \citep{Kurgan2006} identified and reviewed the existing models for the KA process, where the models assumed a number of discrete activities or stages. They identified models consisting of 3 to 9 stages, and commented on the lack of guidance in defining the knowledge discovery process.

However, a certain consensus exists on the major stages of a KA process, regardless of the purpose. A majority of the studies assumes a ``big 3 model'' in support for the introduction of various KA methods. In \citep{Boose1989a} the three stages in KA are data elicitation from a human expert, data interpretation (which infers the underlying knowledge from the data) and knowledge modelling (which transfers the domain knowledge of the human expert into a computational model). In \citep{Cooke1994} KA is considered by Cooke as the front-end of knowledge engineering, which in turn is seen as part of the process of building a knowledge-based system in general. The author considers that the KA process consists of three major steps: knowledge elicitation, knowledge explication, and knowledge formalisation. She states that the final goal of the knowledge acquisition exercise is ``to externalise the knowledge into a form that can be later implemented in a computer'', or in other words the creation of computational models; this idea is also present in \citep{Boose1989a}, \citep{Klein2000} and \citep{Gray2008}. This is achieved through the three stages of the process, where the elicitation provides the psychological data from human experts, the explication categorises and sorts those data and the formalisation transforms them into a computational model. The author also makes a distinction in the first stage between the knowledge elicited from humans and that extracted from other sources \citep{Cullen1988,Fan2012} such as task documentation, historical data, procedure manuals, etc. However, Cooke notes that the knowledge resultant from this extraction process should be verified, and enriched with that coming from elicitation in human SMEs. A decade later, the KA process is presented in \citep{Crandall2006} by Crandall et al. in a similar manner, with the three stages named knowledge elicitation, data analysis and knowledge representation. The authors enrich the description of the stages and also slightly shift their understanding. The elicitation stage is no longer differentiated in elicitation from humans and extraction from alternative sources, and is seen as the activity of collecting information about ``judgements, strategies, knowledge, and skills that underlie performance''. The view of data analysis stage is similar to that of Cooke - it encompasses the set of actions taken for ``structuring data, identifying findings, and discovering meaning''. However, the knowledge representation stage is focused on externalising and presenting the data, which in Cooke's study was the goal of the whole knowledge acquisition process. In Cooke's study, the formalisation stage is explicitly associated with the formulation of a computational model, while Crandall and colleagues state that representation means ``displaying data, presenting findings, and communicating meaning'' that is, communicating the output of the analysis step. In a different approach to the ones described above,  Yates and colleagues \citep{Yates2007,Yates2011} acknowledge the three stages in the KA process; however, they note that the last two, analysis and representation, are usually treated together, being inseparable in the context of modelling the knowledge acquired through elicitation. Researchers in Knowledge Discovery also acknowledge the three stages model, with the three stages named: data preprocessing, data mining, and knowledge interpretation and evaluation \citep{Cano2003,Ngai2009,Mukhopadhyay2014,Mukhopadhyay2014a}. The above discussion is summarised in Table~\ref{Table:KAprocess}, which shows the main studies supporting the three stage model of the KA process, from various perspectives.

\begin{figure*}[!t]
	\begin{center}
		\includegraphics[width=0.8\linewidth]{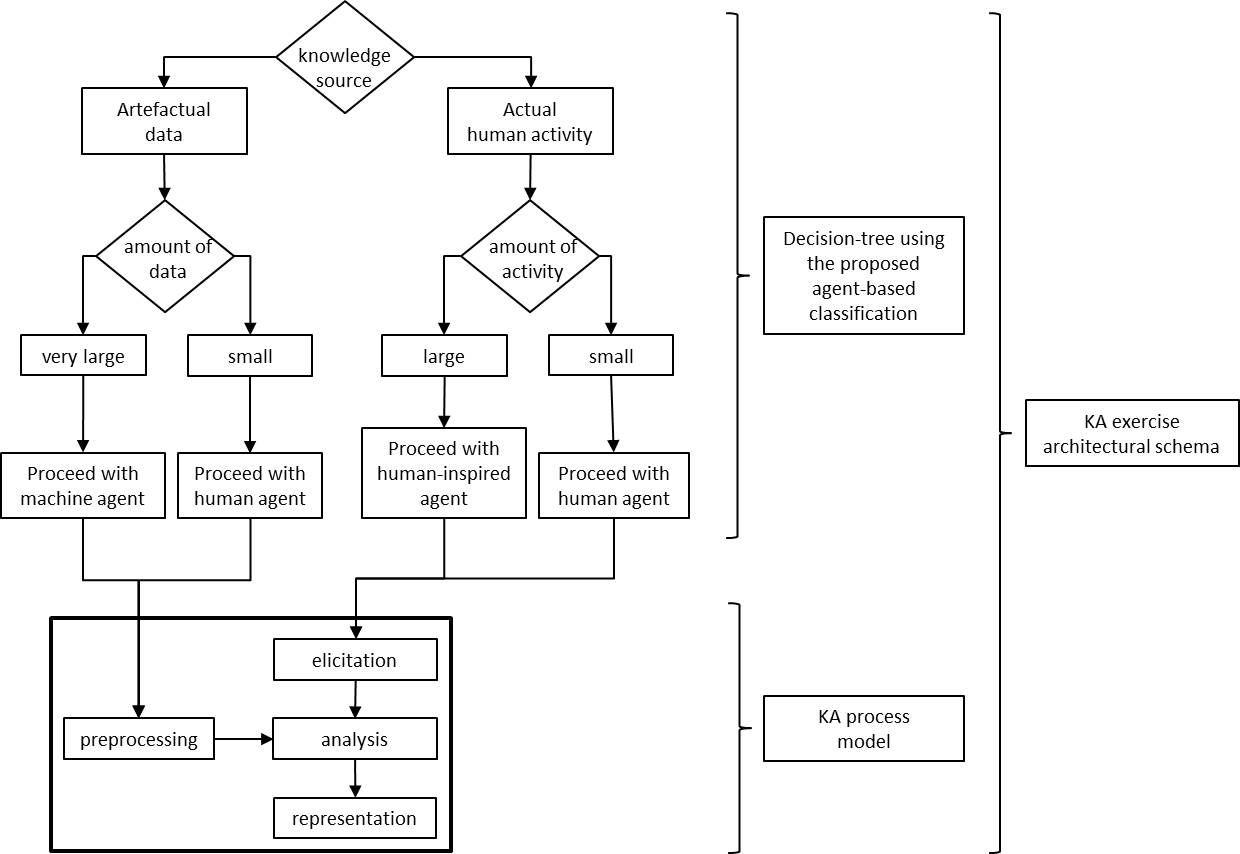}
		\caption{The KA excercise - architectural schema}
		\label{Fig:KAschema}
	\end{center}
\end{figure*}

In general, the need of a universal process model to support the KA exercises is obvious, and we showed in this section that there is no lack of preoccupation in the scientific world for investigating in this direction. In particular for this paper, it was necessary to summarise the existing work and to commit to a process model in order to highlight the relation between the proposed classification and the KA process. Assume that one must gain knowledge about how and why aircraft pilots perform various actions, in order to communicate this knowledge to pilot schools for purposes such as improving training programs. The first question is if the pilots are available in person for the KA exercise or the exercise is performed on artefacts of their activity, such as flight data from the on-board computers. If the pilots are available then it is important to establish how many of them will be participating in the KA exercise. In the case of one or few pilots, a team of human elicitors (i.e. human agency) is sufficient for performing the KA exercise, but if a large number of pilots needs to be involved, such as the whole population of pilots of an airline, then automated/computerised versions of human elicitors (i.e. human-inspired agency) may be needed in order to deal with the amount of data and analysis required. If on the other hand only artefactual data of pilot's activity is available then again the amount of the available data triggers further choice options. If these data are in a low amount, such as simple tables with only the very important decisions like change of course, these can be analysed by humans, but if these data mean detailed records of all maneuvers, then machine agents should be chosen in order to mine the vast amounts of resultant data. In Figure~\ref{Fig:KAschema} we generalise this example and present in a visual manner how the proposed classification can contribute to a KA exercise by creating a preliminary decision tree that facilitates the choice of the appropriate methods to be used within the KA process. Thus, another merit of the agent perspective on KA is that it plays a decision support role in the KA exercises.
 
\begin{table*}[!t]
	\centering
%	\begin{scriptsize}
	%\resizebox{\textwidth}{!}{%
		\begin{tabulary}{\linewidth}{|c|c|c|c|c|}
			\hline
			\multicolumn{1}{|c|}{\textbf{Perspective}} & \multicolumn{3}{c|}{\textbf{Process stages}} & \multicolumn{1}{c|}{\textbf{References}} \\ \hline
			KA & data elicitation & data interpretation & knowledge modelling & \citep{Boose1989a} \\ \hline
			KA & knowledge elicitation & knowledge explication & knowledge formalisation & \citep{Cooke1994} \\ \hline
			CTA & knowledge elicitation & data analysis & knowledge representation & \citep{Crandall2006}, \citep{Yates2007} \\ \hline
			KD & data preprocessing & data mining & knowledge interpretation & \citep{Cano2003}, \citep{Ngai2009}, \citep{Mukhopadhyay2014} \\ \hline
		\end{tabulary}
	%}
%	\end{scriptsize}
	\caption{The three stage KA process model}
	\label{Table:KAprocess}
\end{table*}

\section{Brief review of the existing KA classifications}

As discussed in previous sections, KA is used in a very broad range of research fields and practitioner activities, and while having a fairly short history as a standalone concept, it includes numerous precursor techniques inherited from previous theoretical or practical approaches. This generated diverse paths of development for KA, which further generated numerous approaches customised for every field of research, purpose of the exercise, and even problem or task. This makes KA a highly problem-dependent concept, whose classification into a comprehensive taxonomy was very difficult. In \citep{Roth2014} the authors note that while the need of conducting KA exercises is well established and acknowledged across many fields and problems, there is not much guidance about how to conduct it or evaluate its quality. The number of KA methods identified in the literature over the years is in the range of hundreds, as reported by some of the reviews of the field \citep{Cooke1994,Schraagen2000,Wei2004,Hoffman2008} which attempted to provide the basis for classifications of KA techniques. In the following, we provide a brief description of the most important classifications proposed over time in relation to the KA paradigm.

One of the first notable attempts to categorise KA methods belongs to Bainbridge \citep{Bainbridge1979}, where the author proposed a matrix relating the acquisition method to the type of knowledge/information needed to be acquired. She introduced seven categories of ``desired information'': general information on the effect of  variables, general information on control strategy, numeric information on control strategy, the process, decision sequences, general types of cognitive processes and full range of behaviours. These were linked to acquisition methods such as observation and interviews, verbal protocol, questionnaires, etc. Later, Yates \citep{Yates2007} notes that Bainbridge was also the first to suggest that a combination of techniques must be used in the KA process in order to ensure validity of the exercise; this became a common practice in KA, with a majority of the researchers in the field supporting the idea \cite{Cooke1994,Militello1998,Schraagen2000,Yusoff2012}.

From a different perspective Embrey \citep{Embrey2000} divided the KA techniques into two major categories: action oriented and cognition oriented. The action oriented techniques concentrate on describing and structuring the observed behaviours and tasks, while the cognition oriented techniques focus on the cognitive processes underpinning the observable behaviour. We add that this classification resembles the historical evolution of KA, which we briefly presented earlier in the paper, where in the research on human activity the initial work on studying the tasks was gradually enriched through the inclusion of cognition in the analysis process.

Another classification of KA techniques was based on the type of knowledge acquired by the KA exercise \citep{Cooke1994,Essens1995,Yates2011}. This classification uses the ``differential access hypothesis'' \citep{Hoffman1992}, that is, the belief that different acquisition methods uncover different types of knowledge. Most studies consider two categories of knowledge in this classification: declarative and procedural \citep{Yates2007,Clark2008}. In \citep{Clark2008} the authors state that complex human activity requires the seamlessly integrated use of both declarative (conscious, deliberative) and procedural (unconscious, automatic) knowledge. Thus, the resultant KA research focuses on either or both aspects, depending on the purpose of KA exercise. In a different view, substantive and strategic knowledge terms were first proposed by Gruber \citep{Gruber1989,Gruber1989a} and later discussed in \citep{Essens1995} and \citep{Yates2007}. The substantive knowledge is related to analytic activity, being used for describing and drawing conclusions about the domain of activity and/or state of the world, while the strategic knowledge is used at the decisional stage for deciding what action to perform next, based on the perceived state of the world. If we use the same aircraft pilot example we used earlier in the paper, a pilot uses substantive knowledge to conclude that the flight is in a high risk situation based on the weather data showing a potential storm, and uses strategic knowledge for deciding to descend in order to avoid the turbulent area.

A different taxonomy is based on the stage in the KA process in which the method is performed, where each stage is associated with an output of the acquisition exercise \citep{Olson1991}, e.g. data acquisition, analysis, representation. This categorisation is also indirectly presented by Cook in \citep{Cooke1994}, even though the study is focused on classifying the internal mechanisms of the methods.

Recently, Yates and colleagues \citep{Yates2007,Yates2011} proposed a taxonomy of KA methods which shows how different categories of purposes require the use of specific categories of methods. In essence, the authors assume that in practice the methods are not used individually, but rather in pairs of elicitation methods and analysis/representation methods coupled with specific purposes/activity types.

The most important, and perhaps most recognised and well-established classification of KA methods is based on the internal mechanism of the methods (the mechanism-based classification), that is, the way they acquire the knowledge. A classification that is generally accepted and largely used in the field \citep{Crandall2006,Yates2007,Clark2008,Hoffman2008,Yates2011,Tofel-Grehl2013} assumes four broad families of KA methods: informal techniques (observation and interviews), process tracing techniques, conceptual techniques, and formal models. The first three were proposed by Cooke \citep{Cooke1994}, while the fourth is a later addition by Wei and Salvendy \citep{Wei2004}.

\begin{figure*}[!t] %h]
\begin{center}
    \includegraphics[width=0.75\linewidth]{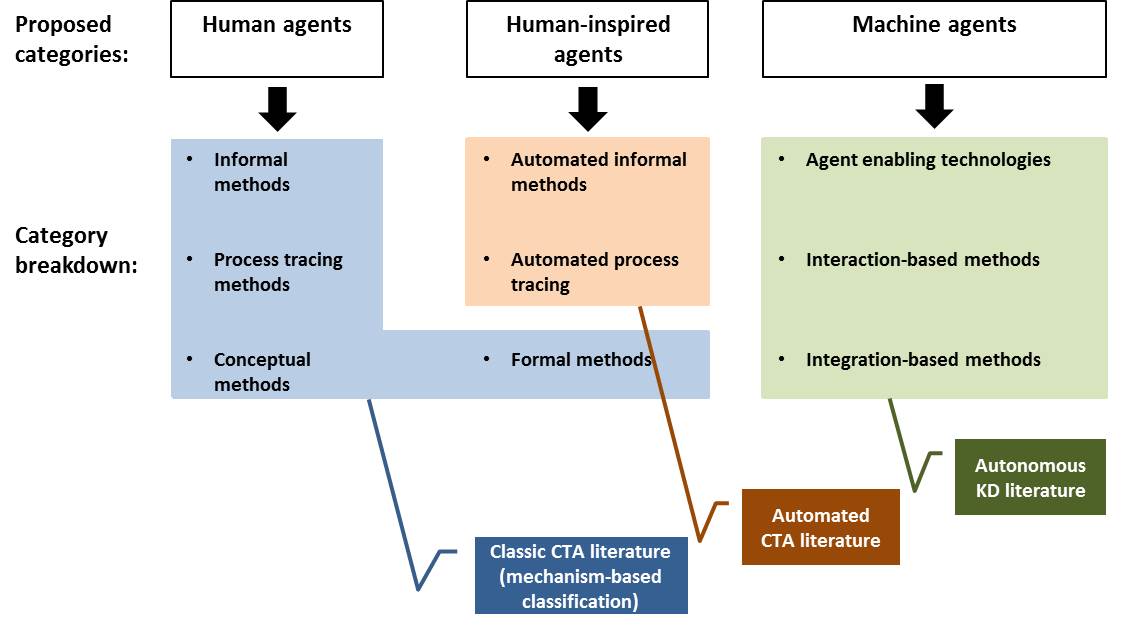}
    \caption{The proposed classification and its relation to the existing mechanism-based classification.}
    \label{Fig:classif}
\end{center}
\end{figure*}

As we stated earlier in the paper, in this study we classify the KA methods from the point of view of the type of agency involved in the KA exercise. However, in each of the categories belonging to the proposed classification we discuss the methods based on the way they implement the KA exercise, that is, based on the internal mechanism of the methods. Consequently, we will also use in our review the mechanism-based classification, and will redistribute the methods from this classification into our classification, accordingly. This redistribution is presented visually  in Figure~\ref{Fig:classif} in support of the next sections. The diagram summarises the main categories in the proposed KA method classification and shows how the subcategories discussed for each category fall into the existing KA literature.

\section{The human agents}

\begin{figure}[h]
\begin{center}
    \includegraphics[width=0.99\linewidth]{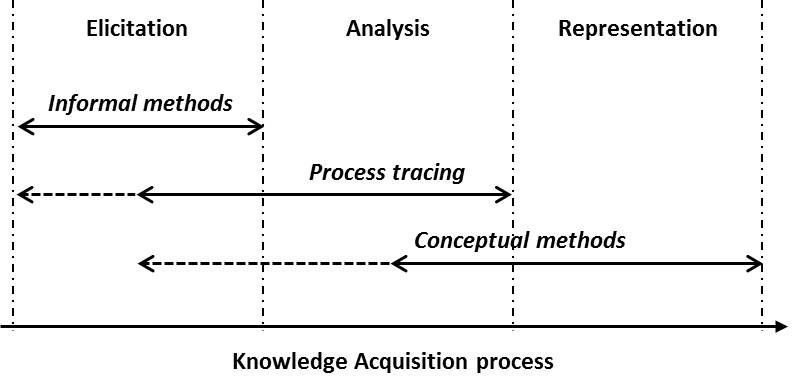}
    \caption{Mapping of he human agent methods to the KA core process: solid line - ample explicit coverage, dashed line - limited implied coverage.}
    \label{Fig:HumanToProcess}
\end{center}
\end{figure}

The knowledge acquisition using human agents has been extensively covered in the literature, with numerous stand-alone methodologies, methods and techniques proposed in the last decades \citep{Cooke1994,Clark2008}. Very well documented reviews of these techniques have been published over the years, especially during the 1990's and early 2000's \citep{Boose1989a,Cooke1994,Schraagen2000,Wei2004}, however, several important reviews which include aspects related to more recent human agent techniques have been also published in the last decade \citep{Yates2007,Clark2008,Hoffman2008,Tofel-Grehl2013}. From the point of view of the classification we propose in this paper, the human agent methods fall in general into the first three families of the mechanism-based classification \citep{Cooke1994}: informal methods, process tracing methods and conceptual methods. In this section, we provide for each of these categories a summary description of the category and a brief presentation of the most important methods belonging to it. Also, in order to guide the review of the literature on human agency in KA, in Figure~\ref{Fig:HumanToProcess} we provide a mapping of the human agent methods onto the KA process.

\subsection{Informal methods - observations and interviews}

Informal techniques are usually methods used for direct acquisition, in which a human elicitor watches human experts and/or talks with them in order to extract knowledge about task performance either from their observed actions, or from their active response to tasks-related enquiries. Methods in this category are highly informal, and are highly verbal and descriptive, requiring the elicitor to either possess certain levels of expertise in the task that is under assessment, or otherwise to be highly qualified and experienced in interviewing techniques. These methods are useful in the initial stage (elicitation) of the knowledge acquisition exercise (Figure~\ref{Fig:HumanToProcess}) for forming an initial view on the problem domain, however, the results may be difficult to interpret.

Observation methods involve extracting features from the observed human behaviour occurring during the performance of a task. The observation can be applied to the actual tasks or to records (video, audio, etc.) of the task performance \citep{Drury1990,Hoffman1987}. In \citep{Meyer1992} the author states that observation is used in ``manual'' knowledge acquisition to familiarise the knowledge engineer with the domain. This familiarisation takes place before other techniques are applied, such as interviews or more formal methods, and facilitates further construction of knowledge models and representation. Observation is used to gather information about conscious and unconscious behaviour of subjects and to investigate the processes involved in the development of expertise. Cooke \citep{Cooke1994} identifies three observation methods: the active participation, the focused observation and the structured observation. While observation is by nature seen as a passive elicitation procedure, the active participation method assumes that the elicitor is involved together with the subjects in task performance. Hoffman \citep{Hoffman1987} also provides a list of task types that can be subject to observations. He describes six families of tasks: familiar tasks, simulated familiar tasks, limited information tasks, constraint processing tasks, combined constraints tasks, and tough cases. In the elicitation process the subject is gradually exposed to situations belonging to various levels of familiarity and difficulty in relation to the domain of interest for the elicitor.

Interviews are another major type of informal human agent methods. They classify in unstructured and structured and can be direct or indirect, with questions that are explicit or implied \citep{Hoffman1987,Cooke1994,Wei2004}. In general they are retrospective - subjects are asked to retrieve information about certain tasks they performed in the past. The unstructured interviews are suitable for early stages of knowledge elicitation \citep{Cooke1994} due to the fact they do not require domain knowledge for the elicitor. The interviews do not follow a predetermined sequence or topic; the elicitor proceeds freely, and uncovers whatever information the subject can recall. For this reason they seem to be the preferred methods for initial stages of knowledge elicitation. However, the drawback is that the elicitor needs previous training and/or skills to facilitate the conversation and guide the interview. In addition, the data provided by unstructured interviews can be ``copious and unwieldy'', as noted in \citep{Wei2004}. The structured interviews are different from unstructured ones through that they follow a process that is in some extent predefined in terms of content and sequencing \citep{Cooke1994}. Depending on how strictly the process follows the predetermined format, they can be structured, semi-structured or prompted. The advantage of structured interviews over the unstructured ones is that they provide systematic, and thus, more complete views of the domains of interest. Also, the structured process tends to take less time, and the participants, both the elicitors and the experts tend to be more comfortable throughout the elicitation exercise. However, structured interviews require more preparation time before the interview actually takes place. Nine types of structured interviews have been identified in the literature \citep{Meyer1992,Cooke1994,Clark2008}, as follows: focused discussions, teach-back, role play, twenty-questions, Cloze experiment, Likert scale items, question answering protocols, questionnaires and group interview techniques.

Starting from these broad families of techniques, in the following we describe the most popular methods, which generated the most important body of research to date.

Critical Decision Method \citep{Klein1989,Wei2004} was proposed in the 1980's and is based on semi-structured interviews, using incidents as facilitators for testing decisions, judgements and general problem solving. The method is time consuming and requires elicitors to possess high level of expertise in the domain of interest, involving multiple-pass retrospection of past events guided by probe questions. However, the method generates information that is very rich and very specific in the same time. It is considered a method for high level KA, and was extensively used by both researchers and practitioners. Reviews of the method and its applications from various historical periods can be found in \citep{Cooke1994,Hoffman1998,Crandall2006,Gazarian2013}. Recently, the method has been extensively applied especially in psychology and general medical practice \citep{Pauley2011,Geis2013,Pauley2013,Steidtmann2013,Wannheden2013,McNelis2014}, but also in ergonomics and human factors \citep{Plant2013,Plant2014} or general critical decision-making problems \citep{Okoli2013,Ahn2014}.

PARI (Precursor-Action-Result-Interpretation) \cite{Hall1994,Hall1995,Wei2004} is a nine stages structured interview methodology which elicits detailed information about procedural skills. The method was proposed in the 1980's and uses multiple experts on a set of problems, being considered to be time consuming and not feasible for usual operational contexts. However, it is considered a highly effective method which provides very detailed analysis of human skills. A simplified version of PARI (S-PARI) reduces the number of stages of the standard PARI from nine to three, and uses pairs of experts in relation to a specific problem. The method breaks down action, precursor and interpretation data into cognitive processes and investigates subtasks on only one task specific problem. The method is considered to be highly effective in analysing procedural skills, and easily usable in operational settings. Reviews of the PARI methodology and its applications from various historical periods can be found in \citep{Cooke1994,Wei2004,Tofel-Grehl2013}, the literature showing that the most important body of research related to PARI methodology was concentrated in the 1990's and early 2000's. However, several important studies using PARI have been also reported in the recent years in instructional design in domains such as healthcare or engineering \citep{Park2012,Clark2014} or in virtual environments applications \citep{Code2012}.

The Task Diagram approach \citep{Klein1998,Wei2004} combines the diagram-based focused discussions with structured interviews and is used for generating task diagrams when a roadmap for task analysis is not in place for the task of interest. The Task Diagram techniques generate an initial overview of the task, providing support for the rest of the cognitive task analysis; however, they have the disadvantage that they cannot handle very complex tasks. Comprehensive reviews of the major body of TD techniques and their applications can be found in \citep{Cooke1994,Schraagen2000,Wei2004,Hoffman2008}. Recently, task diagram methods have been used extensively in various fields, such as in medical domain in clinical research and pathology \citep{Yagahara2013,Wannheden2013}, in serious games \citep{Antonova2011,Seager2011}, in ergonomics and human factors \citep{Naweed2014} or in instructional design \citep{Yusoff2012} applications. In the last decade the classic task diagram methods also evolved towards a different elicitation approach called Graphic Elicitation \citep{Varga-Atkins2009} in which diagrams, photos or various forms of visual arts are used for stimulating experts' thoughts on task structure and performance, or in the reverse process experts are asked to provide visual representation of the personal understanding of tasks, concepts, experiences and behaviours. The literature on Graphic Elicitation reported a substantial number of studies in the recent years, in a variety of application fields \citep{Crilly2006,Driessnack2012,Kuehne2013,Notermans2013,Richard2015}.

The Knowledge Audit \citep{Klein1998,Wei2004} is a structured interview method used in general when the elicitor needs to identify quickly the essential cognitive aspects associated with the task of interest. The method uses six compulsory probes related to past and future, big picture, noticing, task tricks, opportunities for improvement and expert-novice differences, and three optional probes related to anomalies in the task, equipment challenges and tough scenarios. The method is thought to be eliciting very detailed and specific information about the task performance, however the outcome is subjective and hence, may not be very accurate. The method was proposed in the 1980's and a number of reviews on knowledge acquisition describing various knowledge audit techniques and their applications are available in the literature \citep{Cooke1994,Schraagen2000,Cheung2007,Hoffman2008}. Recently the knowledge audit methods have been used in various fields such as serious games \citep{Antonova2011,Seager2011}, business and organisational development \citep{Rahman2011,Gourova2012,Daghfous2013}, transportation and energy sectors \citep{Cheung2007,Ragsdell2014}, and learning and instructional design \citep{Yusoff2012}.

The Simulation Interview method \citep{Klein1998} is used for understanding the task process in which the elicitor brings out the major events throughout the task performance for highlighting the main cognitive elements of the task in relation to a specific incident. The method shows how the subjects perform the task, and also how they think about it, thus, it can deal with both experts and novices. Detailed description, historical notes and reviews on the simulation interview techniques can be found in \citep{Cooke1994,Wei2004}. In recent years simulation interviews have been used especially as part of ACTA (the Applied Cognitive Task Analysis methodology) \citep{Militello1998}, in fields such as clinical research \citep{Pugh2011}, welfare \citep{Bogo2014,Brubacher2015,Powell2015} or ergonomics and human factor \citep{Davidsson2014}.

Another approach largely used in the classic human agent KA exercises is the Cognitive Demands Table method, which synthesises data from multiple interviews, finds patterns and organises them taking into account difficult elements, common errors, cues and strategies from SMEs' behaviour \citep{Klein1998,Wei2004}. The method identifies the components of a task that are cognitively demanding and facilitates their further application in the task design cycles. Versions of the Cognitive Demands Table have been also used as part of ACTA \citep{Militello1998}, especially in very recent studies providing extensive coverage of a wide range of applications in numerous fields, such as medicine, ergonomics and human factors, psychology, economics, learning and instructional design, or games and virtual environments \citep{Pugh2011,Gourova2012,Yusoff2012,Daghfous2013,Bogo2014,Davidsson2014,Brubacher2015,Powell2015}.

A recently emerging approach, thoroughly reviewed in \citep{Chen2012}, applicable in conjunction with any of the methods and family of methods described above, is the Elicitation by Critiquing, in which third party participants in the elicitation process are asked to critique SMEs actions or the resultant products of their actions in order to generate meaningful knowledge about the tasks. The approach have been used especially, but not exclusively, in studies related to recommender system design \citep{Pu2011,Yoo2013,Jansson2015}.

\subsection{Process tracing techniques}

Process tracing techniques typically capture an expert's performance of a specific task through a think-aloud protocol during the performance of the task, or through a subsequent recall of it. Similar to the informal methods, these methods are also highly customised and task-dependant, requiring the elicitor to possess certain expertise in the problem domain. However, unlike the informal methods, they do not necessarily require direct interaction (i.e. subjects can be recorded), and the data to be monitored and extracted from the expert are predefined. Process tracing techniques have a slightly higher degree of formalisation compared to observations and interviews, and can be used in either or both the elicitation and the analysis stages of knowledge acquisition (Figure~\ref{Fig:HumanToProcess}), allowing the exploration of the cognitive structure of task performance. In the literature most of the classic KA studies consider that the process tracing approaches can be split into five major approaches: verbal reports, non-verbal reports, protocol analysis, decision analysis, and cognitive walk-through \citep{Cooke1994,Wei2004}. Several detailed descriptions and historical notes on process tracing can be found in \citep{Cooke1994,Wei2004,Hoffman2008}, while a very recent, thorough and comprehensive review can be found in \citep{Beach2013}. In the following we briefly describe the five main categories and provide references to their most recent applications.

The Verbal Report methods assume that the elicitors extract knowledge from SMEs based on what the latter are able to express through verbal description of what they believe they do for accomplishing the tasks of interest. The verbal communication can be on-line, if the expert provides a verbal report in real-time while performing the task, or off-line if the expert comments retrospectively on how they performed the task in the past. It can be also self-reported, when the expert reports own experience, or shadowed, when a second expert reports the facts experienced by the expert performing the task. Verbal reports are used in the initial stage of knowledge elicitation, and provide the raw data which is further used for protocol analysis. Recent studies on verbal reports have been reported especially in psychology related applications \citep{Fox2011,Lemke2012,Guhde2014}.

The Non-Verbal Report methods \citep{Cooke1994,Fox2011,Salmon2012,Govaerts2013} assume the collection of data other then verbal, for process tracing purposes. Examples of non-verbal data can be bodily communication (e.g. eye movement, facial expression), inputs for computer-based tasks (e.g. keystrokes, mouse, touch screen), or physiological data from sources such as EEG (electroencephalography), EOG (electrooculography), EMG (electromyography), X-Ray etc. Non-verbal reports are also used in the initial stage of knowledge elicitation, like their verbal counterparts, especially when the verbal communication of the process involved in task performance is impaired due to various reasons. The data provided by non-verbal reports is then further used for protocol analysis \citep{Cooke1994}.

The Protocol Analysis methods \citep{Cooke1994,Clark2008} emerged from the need of improving the consistence of data resulted from knowledge elicitation process. Informal methods, as well as the verbal and non-verbal reports are in general time-consuming and generate large amounts of data that are qualitative in nature, complex and unordered, which are usually interpreted with a certain amount of subjectivity by the elicitors, depending on the methods and/or their skills. Protocol analysis methods are in general ways of organising large amounts of so-called ``open-ended'' \citep{Cooke1994} material through objective and systematic identification of specific characteristics. Various techniques have been proposed over time, since the approach emerged in the 1970's, such as content analysis \citep{Bainbridge1979}, which searches for predefined patterns as part of the hypothesis testing process, interaction analysis \citep{Olson1991}, which parses the recorded interaction between the elicitor and the expert for identifying patterns in the expert's  statements, or grounded theory \citep{Pidgeon1991}, which is similar to content analysis except the patterns are not predetermined. Recent applications in various fields of the protocol analysis methods can be found in \citep{Ward2011,Li2012,Perry2013}.

The Decision Analysis methods emerged in the 1980's \citep{Bradshaw1990,Cooke1994,Wei2004} and are used for producing quantitative analysis of decision points within a task, when such points are previously identified using verbal and non-verbal protocols or protocol analysis. Thus, these methods are applied in the analysis stage of the KA process, after the initial raw data extraction. Decision analysis uses formal statistical methods and probability theory in order to provide quantitative information and prediction in relation to the decision points of interest. Recent work on decision analysis can be found in \citep{Durbach2012,Keeney2013}, including a comprehensive review of the period between 2000 and 2011 in \citep{Huang2011}.

Another category of techniques, is the Cognitive Walk-through \citep{Wei2004,Mahatody2010,David2014,Kushniruk2015} in which one or more elicitors work through the task of interest ask a set of questions from the perspective of the user. The procedure involves a task analysis that establishes the sequence of steps a human subject needs to take for performing the task, and the system responses to subject's actions. This is followed by the actual questioning sequence, when the elicitor goes through the task steps asking questions to itself at each step. The data obtained through the walk-through exercise are also in the form of verbal protocols, and can be further analysed with one of the protocol analysis methods. However, the method is an exploratory process and is believed to be easy to implement, having low time and cost requirements \citep{Wei2004}.

\subsection{Conceptual Techniques}

In contrast to the first two categories, conceptual techniques produce structured, interrelated representations of relevant concepts within a problem domain. They are mostly indirect, requiring less or no introspection and verbalisation, and can use multiple sources of problem domain expertise, such as multiple human experts and/or task documentation and logs, historical data, practitioner's literature etc. The output of these sources can be aggregated in order to generate a composite structural representation of the acquired knowledge (Figure~\ref{Fig:HumanToProcess}).

Several broad families of methods are described in the literature in relation to conceptual methods \citep{Cooke1994,Sowa2014}. Concept Elicitation methods focus on establishing a set of key concepts that are essential for understanding the domain/task of interest. These concepts and their interaction can be inferred using various informal techniques, such as structured interviews or focused discussions, if the techniques are adapted for elicitation of essential elements of the domain, rather than the extraction of general data. Cooke \citep{Cooke1994} identifies structured interviews as the most appropriate to be used for concept elicitation. The interviews are adapted to concept elicitation through that the experts/subjects are guided towards building dictionaries of concepts related to the task of interest, in which the concepts are grouped based on a set of criteria relevant to the associated domain. The concept elicitation methods can be also based on goal decomposition, involving the construction of hierarchies of concepts which generate taxonomies of concepts. Another notable family of conceptual methods is the Data Collection, which involves the estimation of the degree of correlation between two concepts belonging to a domain of interest. In \citep{Cooke1994} the author refers to this correlation as ``relatedness'' or ``proximity''. These methods are used in general after the initial elicitation of concepts, in conjunction with informal methods or through process tracing, and constitute the immediate support for representation stage of the KA process. The data collection techniques generate one or several matrices of proximity, depending on the number of experts involved in the elicitation exercise, in which the rows and columns represent various domain concepts. Methods in this family include ``rating and ranking'' \citep{Davis2006,Harzing2009}, repertory grid \citep{Bradshaw1993}, sorting \citep{Rugg1992,Davis2006}, event co-occurence \citep{Cooke1994}, and correlation/covariance \citep{Cooke1994}. Another important group of methods is gathered in the Structural Analysis family, where the techniques are based on descriptive multi-variative statistics \citep{Cooke1994} and are in general used for reducing the taxonomies of concepts and the subsequent relatedness estimate matrices to simpler forms which are more appropriate for further knowledge representation. Important methods for structural analysis are multidimensional scaling \citep{Borg2005}, discrete techniques (e.g. clustering) \citep{Jain1999}, direct structure elicitation \citep{Sowa2014}, and structure interpretation \citep{Olson1991}.

In the following we describe a number of conceptual techniques considered to be of major importance in a variety of fields of research from a human agent KA perspective.

The Conceptual Graph Analysis \citep{Gordon1993} method creates a conceptual map of the task in the form of graphs, in which nodes represent conceptualised states, events, goals or actions, and directional links represent relations between the concepts captured in nodes. The graphs can represent taxonomic, spatial and causal structures of the concepts and create the support for knowledge representation through information integration and organisation. The method represents a structured framework for transferring knowledge from an implicit form to an explicit one.

The Consistent Component methods \citep{Fisk1988,Wei2004} investigate procedural knowledge and are used in the intermediate or late stages of analysis. They decompose the task of interest and identify decision points and the automated skills associated to them, providing in the same time measurements for the degree of interference between simultaneous tasks and detailed information about action and skill performance timing \citep{Ryder1993}.

The Diagramming methods are knowledge representation techniques which show the key concepts that tie all the other concepts related to the task/domain of interest \citep{Reed2004,Wei2004}. Diagrams are very intuitive  and in general require low time and cost, being simple ways to represent fairly complex tasks. However, above a certain level of task complexity the diagrams themselves may become too complex to provide good representation of the domain, and require the analysts to employ other methods in order to produce better representations.

The Error Analysis methods \citep{Wei2004} focus on identifying error sources and on classifying these errors. The errors made by subjects throughout the task performance process are systematically analysed in order to establish their relationship with the cognitive processing, i.e. they are mapped to the corresponding cognitive processing failures. Error analysis produces in-depth insights into the intimate cognitive processes and functions and create the premises for realistic representation of thought processes. However, their use is in general appropriate for tasks which are susceptible to errors, such as critical incidents or safety critical domains \citep{Kim2001}.

The Psychological Scaling methods belong to the rating and ranking class described in \citep{Cooke1994}, and focus on the relations between concepts related to a task/domain of interest, with inclusion of subjective aspects such as individual preference and perception. The relations are expressed using estimated proximity between concepts, which are mapped into proximity matrices. The concepts are ranked by using each of them at a time as reference and measuring the similarity with the remaining ones \citep{Ryder1991,Wei2004}.

The Paired Comparison methods are also part of the rating and ranking class, and involve the rating of pairs of concepts in relation to a standard pair \citep{Cooke1994}. Based on relatedness/proximity measures, the techniques generate magnitude estimations for each pair; however, if the number of concepts increases \citep{Wei2004} the methods may become time consuming. Another drawback of the methods is that they only consider relatedness along one specified dimension.

The Repertory Grid methods \citep{Bradshaw1993,Wei2004} are particular cases of general rating and ranking methods, in which the concepts related to the task of interest are rated along ``dichotomous dimensions'' called constructs \citep{Cooke1994}. The constructs can be organised in hierarchies and used for separating the domain concepts. The subjects provide to the elicitor ratings of various constructs, which are then placed into grids, on rows and columns. Overall relatedness/proximity is then extracted from the grid, by calculating the correlation between ratings constructs and concepts. The result of repertory grid methods is usually further used for knowledge modelling, through structural modelling techniques such as hierarchical clustering.

The Sensori-Motor Process Charts are techniques that focus on mental activities \citep{Wei2004,Jun2012}. They take into account sensors and sensory information and their relation to the task of interest and produce skill charts based on the patterns observed during the task performance. In general they follow a predefined set of steps, equivalent to the key mental activities involved in psycho-motor tasks: plan, initiate, control, end, and check \citep{Wei2004}.

The Sorting methods focus on high-level conceptual structures \citep{Rugg1992,Cooke1994,Wei2004,Davis2006}. The experts are required to order concepts based on relatedness/proximity and place them into piles, accordingly. In addition, due to the fact the experts are not restricted in terms of number and content of piles (one concept can be placed in multiple piles), these can be also asked to label the piles. In general the sorting techniques are considered to be less time-consuming then the paired comparison, with which they are alike, but instead they have a lower level of sensitivity to relatedness/proximity differences. In general sorting techniques are considered fast and simple ways of eliciting conceptual information of a quality that is good enough to be used reliably in further analysis of data.

\subsection{Summary of human agents}

We included in this section those methods which suited the perspective depicted earlier in the paper in Figure~\ref{Fig:KAschema}. In Table~\ref{Table:summaryHA} we summarise this section in a tabular manner and through this, we emphasise the contribution of this class of methods to the creation of the roadmap that facilitates the KA process, and we show how each of the methods can fill an appropriate branch in the decision tree, corresponding to human agency. Thus, overall, this section contributes to demonstrating the decision support role of the human agent class of methods in the KA exercises.

\begin{table*}[!t]
	\centering
%	\begin{scriptsize}
		\begin{tabulary}{\linewidth}{|c|c|c|c|c|c|c|}
			\hline
%			\multicolumn{1}{|c|}{\textbf{KA}} & \multicolumn{2}{c|}{\textbf{Agent perspective}} & \multicolumn{3}{c|}{\textbf{KA}} & \textbf{References} \\			
%			\multicolumn{1}{|c|}{\textbf{approach}} & \multicolumn{2}{c|}{\textbf{roadmap}} & \multicolumn{3}{c|}{\textbf{process model}} & \\ \cline{2-6} %\hline
			& \multicolumn{2}{c|}{\textbf{Agent perspective}} & \multicolumn{3}{c|}{\textbf{KA}} & \\			
			\textbf{KA} & \multicolumn{2}{c|}{\textbf{roadmap}} & \multicolumn{3}{c|}{\textbf{process model}} & \textbf{References} \\ \cline{2-6} %\hline			

			\textbf{approach} & \multicolumn{1}{c|}{\textbf{Amount of}} & \multicolumn{1}{c|}{\textbf{Richness of}} & \multicolumn{1}{c|}{\textbf{Elicitation}} & \multicolumn{1}{c|}{\textbf{Analysis}} & \multicolumn{1}{c|}{\textbf{Representation}} & \\ 
			
			& \multicolumn{1}{c|}{\textbf{activity}} & \multicolumn{1}{c|}{\textbf{analysis}} & \multicolumn{1}{c|}{} & \multicolumn{1}{c|}{} & \multicolumn{1}{c|}{} & \\ \hline
			
			Critical Decision & very low & very rich & yes & yes & no & \citep{Cooke1994,Hoffman1998,Crandall2006,Gazarian2013}\\ \hline
			PARI & very low & very rich & yes & yes & no & \citep{Cooke1994,Wei2004,Tofel-Grehl2013} \\ \hline
			Task Diagram & very low & moderate & yes & partial & no & \citep{Klein1998,Schraagen2000,Wei2004,Hoffman2008} \\ \hline
			Knowledge Audit & low & rich & yes & partial & no & \citep{Cooke1994,Schraagen2000,Cheung2007,Hoffman2008} \\ \hline
			Simulation Interview & moderate & moderate & yes & partial & no & \citep{Cooke1994,Militello1998,Wei2004} \\ \hline
			Cognitive Demands Table & high & moderate & yes & yes & partial & \citep{Klein1998,Militello1998,Wei2004}\\ \hline
			Elicitation by Critique & very low & moderate & yes & no & no & \citep{Pu2011,Yoo2013,Chen2012,Jansson2015}\\ \hline
			Verbal Reports & very low & low & yes & no & no & \citep{Fox2011,Lemke2012,Beach2013,Guhde2014}\\ \hline
			Non-Verbal Reports & very low & moderate & yes & no & no & \citep{Cooke1994,Fox2011,Salmon2012,Govaerts2013}\\ \hline
			Protocol Analysis & moderate & rich & no & yes & no & \citep{Bainbridge1979,Cooke1994,Clark2008}\\ \hline
			Decision Analysis & moderate & rich & no & yes & no & \citep{Bradshaw1990,Wei2004,Huang2011}\\ \hline
			Cognitive Walk-through & very low & low & yes & yes & no & \citep{Wei2004,Mahatody2010,David2014,Kushniruk2015} \\ \hline
			Conceptual Graph Analysis & high & rich & no & yes & no & \citep{Gordon1993,Cooke1994}\\ \hline
			Consistent Component & moderate & rich & yes & yes & no & \citep{Fisk1988,Ryder1993,Wei2004}\\ \hline
			Diagramming & moderate & rich & no & partial & yes & \citep{Reed2004,Wei2004} \\ \hline
			Error Analysis & low & moderate & yes & yes & no & \citep{Wei2004,Kim2001}\\ \hline
			Psychological Scaling & moderate & rich & no & yes & yes & \citep{Ryder1991,Cooke1994,Wei2004} \\ \hline
			Paired Comparison & moderate & moderate & no & yes & yes & \citep{Cooke1994,Wei2004} \\ \hline
			Repertory Grid & moderate & rich & no & yes & partial & \citep{Bradshaw1993,Wei2004} \\ \hline
			Sensori-Motor Process Charts & low & moderate & yes & yes & partial & \citep{Wei2004,Jun2012} \\ \hline
			Sorting & high & moderate & yes & no & no & \citep{Rugg1992,Cooke1994,Wei2004,Davis2006} \\ \hline			
		\end{tabulary}
%	\end{scriptsize}
	\caption{Summary of human agent methods: agent perspective roadmap and KA process}
	\label{Table:summaryHA}
\end{table*}

\section{The human-inspired agents}

As we explained earlier in the paper, we consider the human-inspired agent methods as those methods that automate some of the human agent methods in order to increase the speed of the exercises or to allow the exercises to be performed in cases where physical presence or participation of human elicitors in the KA exercise is impeded or inefficient/inconvenient due to various reasons. From the point of view of the classification we discuss in this study, the human-inspired agents represent a significant contribution to the KA body of research through that they are a significant step in the process of advancing the KA field from purely human-based activities to purely human-independent activities. While the human-inspired agent methods have been extensively used for a variety of KA exercises, they do not present novel approaches regarding the intrinsic mechanisms of the acquisition, but rather they bring novelty in the way the human-based methods, such as interviews or analysis of verbal protocols, are implemented in computer programs. Thus, the contribution stays in the ability of those programs to accurately emulate and automate the well-known human elicitor actions, and to increase the speed and convenience of the KA exercises.

\begin{figure}[h]
\begin{center}
    \includegraphics[width=0.99\linewidth]{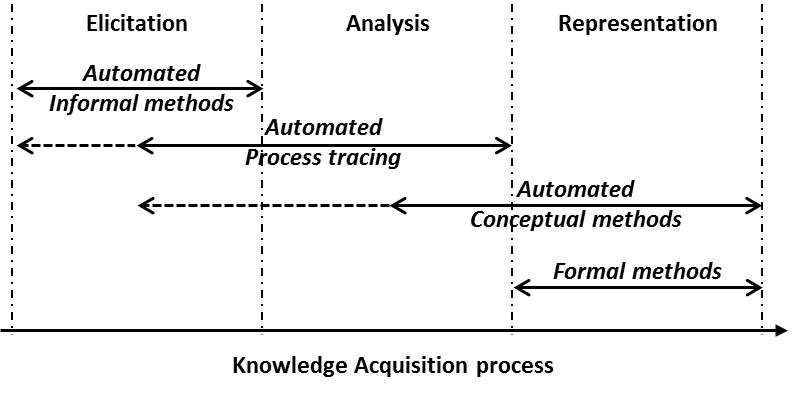}
    \caption{Mapping of he human-inspired agent methods to the KA core process: solid line - ample explicit coverage, dashed line - limited implied coverage.}
    \label{Fig:HumanInspToProcess}
\end{center}
\end{figure}

In this category, the literature contains a variety of human-inspired agent methods which are used for implementing automated versions of some of the informal methods (especially automated interviews), process tracing methods (especially analysis of verbal and non-verbal protocols) and conceptual methods (especially psychological scaling and repertory grid implementations) \citep{Gaines1992,Cooke1994}. However, from the mechanism of the method point of view, most of the human-inspired agents are those belonging to the class of formal KA methods proposed by Wei and Salvendy \citep{Wei2004,Clark2008}. In order to guide the review of the literature on human-inspired agency in KA, in Figure~\ref{Fig:HumanInspToProcess} we provide a mapping of the human-inspired agent methods onto the KA process.

\subsection{Automation of human agent methods}

For the human agent informal methods most of the human-inspired counterparts consist of computer-based tools which include support for fully automated or mixed-initiative (semi-automated) interviews. Some of the most important tools reported in the literature \citep{Cooke1994} are Cognosys \citep{Woodward1990}, which transfers the SME domain knowledge into a graph structure, MORE \citep{Kahn1985} and its enhanced version MOLE \citep{Eshelman1987} which were largely used for generating models of diagnosis knowledge able to disambiguate under-specified domains and to refine incomplete knowledge bases, SALT \citep{Marcus1989}, which was used for problems such as configuration and scheduling for generating expert systems able to handle propose-and-revise problem-solving strategies, and ASK \citep{Gruber1989} which interviews the SMEs for eliciting strategic knowledge about the domain of interest. Other computer based tools containing automated interviews have been also reported in the literature, such as ETS (expertise transfer system) and its enhanced version AQUINAS \citep{Boose1989} which were mainly used by Boeing, IRA-Grid \citep{Linster1993} which acquires descriptions of prototypical situations through a grid-based interview component, or KRIMB (Knowledge Representation for Intelligent Model Building), which acquires descriptive information about complex reliability systems \citep{Diederich1987}.

For human agent process tracing category of techniques, most of the human-inspired counterparts concentrate on implementing computer-based verbal protocol analysis methods in order to automatically record and analyse transcripts from SMEs thinking aloud about tasks. Popular tools reported in the literature were Cognosys, which contained a protocol analysis component in addition to the automated interview capabilities, and KRITON \citep{Diederich1987} which transforms verbal protocols into propositions based on pauses in the speech, along with other tools such as LAPS \citep{Hoffman1995}, MACAO \citep{Aussenac-Gilles1994}, or MEDKAT \citep{Jagannathan1985}.

Conceptual human agent methods were also addressed by various automated tools. Automated counterparts of psychological scaling, including multi-dimensional scaling, were largely used in the literature for structuring knowledge, i.e. AQUINAS, KRITON, IRA-Grid with variations such as FLEXIGrid, KSS0 and others \citep{Boose1989a,Chao1999}. Also a consistent body of research concentrated on repertory grids and used various personal-construct psychology-related methods to elicit and analyse knowledge, such as AQUINAS, ETS and IRA-Grid, FLEXIGrid, KRITON, KSS0, KITTEN, SMEE \citep{Neale1988,Ford1993}.

Recent work that applies the above methods has been reported especially in domains that are new and less explored by the classic KA community, such as games and virtual environments, virtual worlds and societies, or serious games \citep{Bainbridge2007,Petrovic2009,Overbey2010,Hasler2013}, where the knowledge is not acquired (automated or not) from entities that are humans or artefactual data of human activity, but rather from computer generated artificial entities. Such an approach was used in \citep{Overbey2010}, where the authors created automated data collection entities for virtual worlds, which they tested in the Second Life environment. They demonstrated that the entities were capable to capture a variety of behavioural data about various avatars populating the Second Life, data that were further analysed using Social Network Analysis methods. They concluded that a combined approach utilising manual and automated techniques can provide valuable knowledge about elements of interest in virtual worlds and social networks, such as structural and functional aspects, or key players. In a different study Yee and Bailenson \citep{Yee2008} describe a method for collection of longitudinal behavioural data from virtual worlds, and introduce a technical framework for capturing avatar-related data for several weeks in real-time, at a resolution of less than one minute. Also, in \citep{Hasler2013}, the authors used the game Second Life to implement human-inspired agents in embodied survey ``bots'', in the form of software-controlled avatars, in order to obtain automated data collection in 3D virtual worlds. In their framework a human elicitor controlling an avatar and a ``bot'' artificial elicitor controlling another avatar cooperated to perform a survey interview on various characters (both human-controlled avatars and computer generated avatars) populating the game Second Life. They found that both the human-based and the bot interviewer performed the survey in good conditions, however, the bot had a slightly lower response rate compared to the human. While a difference existed, the authors concluded that the findings provided sufficient support for the idea of using bots as virtual research assistants.

\subsection{The formal methods in the mechanism-based KA classification}

The formal methods use computer simulations to model human activity, thus they account for the knowledge representation stage only (Figure~\ref{Fig:HumanInspToProcess}), in the KA process. The acquisition of knowledge does not rely on actual performance of the tasks or records of it, instead is based on cognitive models built on various assumptions which lead to descriptions of the tasks/domains of interest convenient for the purpose of the KA exercise \citep{Wei2004}. The methods in this category are considered by Wei and Salvendy \citep{Wei2004}, and further acknowledged by other authors \citep{Clark2008,Crandall2006,Hoffman2008} as KA methods inspired from assumed understanding of the human mind and cognition, and formalised into computational models that plausibly emulate human activity. We complement this view by saying that the methods in this category consist of attempted cognitive architectures for general intelligence, emerging from the corresponding cognitive theories, and aiming to emulate human problem-solving and decision-making capabilities. Thus, these methods are inherently automated, without human agent counterparts, and therefore, we discuss them in our classification (1) as part of the category of human-inspired agents and (2) separate from the automated versions of the human agent methods. In the following we present the most notable approaches belonging to this category.

The ACT model (Adaptive Character of Thought) comes from the ``ACT*'' cognitive theory initially proposed by Anderson in 1983 \citep{Anderson1983}, and further instantiated as a cognitive agent architecture \citep{Anderson1996,Byrne1997,Salvucci2013}. In ACT the authors consider that cognition emerges from the interplay between procedural and declarative knowledge. Procedural knowledge is modelled through units called production rules encoding transformations in the environment. Declarative knowledge is modelled through units called chunks encoding objects in the environment. The ACT model is focused on skill acquisition and can be applied only to problem solving domain, assuming a ``means-ends'' problem solving structure \citep{Wei2004}. It can represent both declarative and procedural knowledge and represents a way to understand the learning of complex problem solving skills.

The ARK model (ACT-based Representation of Knowledge) \citep{Geiwitz1988,McCloskey1991,Cooke1994,Wei2004} is a technique similar to the goal decomposition, which is part of the informal methods in Cooke's taxonomy \citep{Cooke1994}. The model uses a process that breaks goals into sub-goals and/or actions and generates a network of objects and their interaction by using ACT-inspired production rules for modelling goal-subgoal and goal-action relations. The model is thus able to emulate both the network-based representation of the domain knowledge and the corresponding procedures performed on that knowledge.

The Human Processor model is based on the theory of human information processor \citep{Card1983,Card1986}. It was developed as a model of human-computer interaction, and consists of a series of three processors - perceptual, cognitive and motor - and a set of general purpose memory stores \citep{Liu2006}. From a KA perspective the approach can model various parameters which participate in breaking down of complex tasks into relevant components, taking into account timing attributes. The task decomposition is very detailed, with the associate cognitive processes and skills reaching an ``atomistic'' level \citep{Wei2004}. The model is considered to be very accurate, especially for modelling simple tasks, where detailed task decomposition can be easily done.

GOMS (Goals, Operators, Methods and Selection rules) was also developed as a human-computer interaction model, based on human information processor \citep{Kieras1988,Carroll1988,Elkerton1991,Arend1991,Gray1993}; however, the task modelling and decomposition focuses on higher levels of cognitive processes, i.e. goals, operators, methods and selection. From a KA perspective GOMS is applicable to error-free tasks for which the sequence of actions needed to perform them is known. Based on that particular sequence GOMS is able to provide good understanding of the cognitive interaction between the system user and the system, based on that particular sequence. Consequently, the model is functional for the specificity of the task and needs major rework in order to be ported to different tasks. A number of versions of GOMS model have been proposed over time, such as Basic GOMS, Keystroke Model, Model Unit Task, Natural GOMS Language or Cognitive-Perceptual-Motor GOMS \citep{Wei2004,Stanton2013}.

The Grammar methods \citep{Eberts1997,Wei2004}, such as Cognitive Grammar \citep{Reisner1981} or Task-Action Grammar \citep{Payne1989}, describe the tasks requiring human-computer interaction in a formal language, in which a syntax is associated to the actions taken throughout the task performance. Grammars are considered appropriate for complex systems, and fairly easy ways of modelling languages for human-computer interaction. In Cognitive Grammar methods the grammar is modelled using five types of symbols: terminal - are associated to actions that involve learning and remembering, non-terminal - represent sets of actions that can be grouped based on their similarity, starter - are associated to high-level tasks, meta-symbols - refer to operators/operations such as \textit{and}, \textit{or}, \textit{inclusion}, and rules - which define interactions within the grammar structure. The Task-Action Grammar methods usually identify primitive tasks that do not require control frameworks, and can be performed without involvement of problem solving skills. These tasks are then described in a dictionary using semantic categorisation. Based on the dictionary, rules that associate the simple tasks to actions are generated.

The Object-Oriented Models use object oriented technologies for integrating procedural models, such as GOMS and semantic models, such as grammars \citep{Beringer1991,Wei2004,Stanton2013}. In general it is considered that GOMS-type models are limited to specific types of tasks and subsequent functionality, while semantic models are entirely task-dependant. Object Oriented modelling allows variation of functionality through addition/removal of classes and methods, and due to this flexibility can generate semantic structures for virtually any task. From the KA point of view it can handle both declarative and procedural knowledge, especially in human-computer interaction contexts, and bridges a gap between high-level semantic description of task domain and procedural description of the sequence of actions.

The standalone Cognitive Simulation methods \citep{Roth1992,Wei2004} use inputs from domain scenarios to generate realistic cognitive computational models which can then be compared to observed human behaviour for the same scenarios. In general, such models evolved towards cognitive agent architectures for general intelligence, where the resultant cognitive models were intended to produce behaviour not limited to the same scenarios used for building the models. However, from a KA perspective, the cognitive simulation limits its scope to reproducing, through simulation, plausible behaviours related to the tasks of interest; behaviour that can be be further analysed and decomposed towards increasing levels of detail. CES (Cognitive Environment Simulation) is a Cognitive Simulation approach proposed by Roth and colleagues \citep{Roth1992} for modelling the cognitive activity of operators in nuclear power plants, in order to capture the cognitive demands involved in dynamic fault-management situations. CES monitors and tracks process changes, identifies unexpected behaviours and formulates hypotheses related to these, builds and revises a situation assessment, and formulates proposed actions based this assessment. Other important cognitive simulation approaches such as SOAR \citep{Laird2012}, CLARION \citep{Sun2006}, CogAff \citep{CogAff2013} or SoM (Society of Mind) \citep{Leu2014a} have been also proposed and continuously improved over time, emerging towards complex cognitive architectures largely used nowadays for modelling knowledge and behaviour in a variety of application fields \citep{Leu2014a}.

\subsection{Summary of human-inspired agents}

We included in this section those methods which suited the perspective depicted earlier in the paper in Figure~\ref{Fig:KAschema}. In Table~\ref{Table:summaryHIA} we summarise this section in a tabular manner and through this, we emphasise the contribution of this class of methods to the creation of the roadmap that facilitates the KA process, and we show how each of the methods can fill an appropriate branch in the decision tree, corresponding to human-inspired agency. Thus, overall, this section contributes to demonstrating the decision support role of the human-inspired agent class of methods in the KA exercises.

\begin{table*}[!t]
	\centering
%	\begin{scriptsize}
		\begin{tabulary}{\linewidth}{|c|c|c|c|c|c|c|}
			\hline
			& \multicolumn{2}{c|}{\textbf{Agent perspective}} & \multicolumn{3}{c|}{\textbf{KA}} & \\
			\multicolumn{1}{|c|}{\textbf{KA}} & \multicolumn{2}{c|}{\textbf{roadmap}} & \multicolumn{3}{c|}{\textbf{process model}} & \textbf{References}\\ \cline{2-6} %\hline
			
			\multicolumn{1}{|c|}{\textbf{approach}} & \multicolumn{1}{c|}{\textbf{Amount of}} & \multicolumn{1}{c|}{\textbf{Richness of}} & \multicolumn{1}{c|}{\textbf{Elicitation}} & \multicolumn{1}{c|}{\textbf{Analysis}} & \multicolumn{1}{c|}{\textbf{Representation}} & \\ 
			
			\multicolumn{1}{|c|}{} & \multicolumn{1}{c|}{\textbf{activity}} & \multicolumn{1}{c|}{\textbf{analysis}} & \multicolumn{1}{c|}{} & \multicolumn{1}{c|}{} & \multicolumn{1}{c|}{} & \\ \hline
			
			Automated Informal Methods & high & rich & yes & yes & no &  \citep{Eshelman1987,Diederich1987,Boose1989,Gruber1989} \\
			& & & & & & \citep{Marcus1989,Woodward1990,Linster1993} \\ \hline
			Automated Process Tracing & high & rich & yes & yes & partial & \citep{Jagannathan1985,Diederich1987,Aussenac-Gilles1994,Hoffman1995}\\ \hline
			Automated Conceptual Methods & high & rich & no & yes & partial & \citep{Neale1988,Boose1989a,Ford1993,Chao1999} \\ \hline
			ACT & moderate & moderate & no & no & yes & \citep{Anderson1983,Anderson1996,Byrne1997,Salvucci2013} \\ \hline
			ARK & moderate & moderate & no & no & yes & \citep{Geiwitz1988,McCloskey1991,Cooke1994} \\ \hline
			Human Processor & low & rich & no & no & yes & \citep{Card1983,Card1986,Liu2006} \\ \hline
			GOMS & moderate & moderate & no & no & yes &  \citep{Carroll1988,Elkerton1991,Gray1993,Stanton2013} \\ \hline
			Grammars & moderate & moderate & no & no & yes & \citep{Reisner1981,Payne1989,Eberts1997} \\ \hline
			Object-Oriented Models & moderate & low & no & no & yes & \citep{Beringer1991,Wei2004,Stanton2013} \\ \hline
			Cognitive Simulation & low & rich & no & no & yes & \citep{Sun2006,Laird2012,CogAff2013,Leu2014a} \\ \hline						
		\end{tabulary}
%	\end{scriptsize}
	\caption{Summary of human-inspired agent methods: agent perspective roadmap and KA process}
	\label{Table:summaryHIA}
\end{table*}

\section{The machine agents}

In the previous two categories (human agents and human-inspired agents) the KA relies on human or human-like agents to extract the underpinnings of the observable behaviour of humans. The third category, machine agents, opposes to them in two aspects, as we explained in the introductory section. First, the machine agents acquire knowledge in ways that no longer resemble or depend on the human ways of performing the acquisition tasks. Thus, the acquisition process implemented by the machine agents is not intended to be plausible from the point of view of human elicitor actions, while still delivering consistent results in terms of the KA output. Second, the machine agents no longer acquire knowledge from humans or planned/attended records of their behaviour, but rather from unattended artefactual data resulted from human activity systems. Thus, the machine agents employ a variety of computational intelligence technologies that enable them to perform a human-independent autonomous knowledge discovery process, in order to generate models of the knowledge hidden in the artefactual data. In order to proceed with the survey on the machine agent KA literature, we first describe in a brief manner the most important of these agent-enabling technologies.

\subsection{Agent-enabling technologies}

In this study we consider three classes of computational intelligence methods which are of high importance in enabling the machine agents for KA: statistical analysis, machine learning, and evolutionary computation. The methods in these categories are used either individually or combined in order to implement agents capable to act autonomously for performing the acquisition process.

\subsubsection{Statistical analysis}

Numerous statistical methods have been used over the years for knowledge acquisition, especially in autonomous knowledge discovery contexts, for extracting both the structure \citep{Jonassen2013} and the causal relations \citep{Holl1986} existing in artefactual data. Statistical analysis can be seen in the KA perspective as the mathematical way to capture and disseminate data, with the purpose of defining models for prediction.

Bayesian networks \citep{Neapolitan2012}, rule sets (crisp, rough or fuzzy) \citep{Jagielska1999}, k-means techniques \citep{Kanungo2002}, regression analysis \citep{Draper2014}, or decision tree analysis \citep{Rokach2007} are some of the most popular approaches, which generated a significant body of research in the general KA field, as well as in the KA related to human activity systems, which is of interest in this paper. Comprehensive reviews of these techniques can be found in \citep{Afifi1972} in relation to their mathematical foundation, and in \citep{Benjamini2010,Kantardzic2011} in relation to their application in the KA field.

Methods in the statistical analysis family of agent-enabling technologies benefit from a strong mathematical foundation, through which they can provide well defined and reliable insights into the mechanisms underpinning the artefactual data. However, \citep{Begoli2012} they can only infer the knowledge models from well structured data, while they produce less clear results when dealing with complex and highly non-linear data or multidimensional datasets.

In machine agents for KA, the statistical analysis methods are mainly used as part of the inference modules \citep{Kadhim2014,ChemChem2015}, especially for the early stages in the knowledge discovery process, such as data preprocessing, but also in clustering or rule mining.

\subsubsection{Machine learning technologies}

Machine learning technologies employ a variety of techniques in order to endow artificial entities (machines) with the ability to autonomously learn facts about various phenomena, which from a KA perspective equals to the ability to uncover and understand the inherent structures and causal relations underlying the artefactual data of interest. The machine learning techniques may be supervised, where the supervised learning methods assume the existence of prior (historical) domain knowledge, or unsupervised, where the machine entities act entirely independent in the knowledge discovery process.

Inductive Logic Programming \citep{Corapi2012,Nguyen2013}, Support Vector Machines \citep{Orru2012,Rebentrost2014}, Reinforcement Learning \citep{Lee2012,Kober2012}, and Artificial Neural Networks \citep{Jagielska1999,Carpenter2005,Chattopadhyay2014} are the most popular approaches in machine learning which generated the largest body of research in the KA field. Detailed reviews of the machine learning field and techniques can be found in \citep{Kantardzic2011}, \citep{Mohri2012} and \citep{Marsland2014}.

Often machine learning techniques are used in conjunction with statistical methods for extracting the meaningful information out of the non-linear and multidimensional datasets, i.e. neural networks can be used in conjunction with symbolic production systems in the form of a rule sets, where the rules can be crisp (if-then), rough or fuzzy. Consequently, the machine agent methods for KA can employ one or more machine learning techniques, which in turn may make use of statistical methods, resulting in complex integrated machine agents covering more than one stage/sub-process in the KA process.

\subsubsection{Evolutionary Computation technologies}

The motivation for applying evolutionary computation (EC) techniques to knowledge acquisition is that they are robust and adaptive search techniques that perform a global search in the solution space. Evolutionary computation techniques are not used \textit{per se} as standalone methods to implement acquisition tasks such as feature selection, rule mining, clustering or classification. Rather, they are used to evolve the parameters of other methods in order to improve the quality of the incipient knowledge extracted by those methods. Thus, evolutionary computation techniques have been found particularly useful in automatic processing of large quantities of raw noisy data, where large numbers of parameters used by various other KA techniques needed to be optimally set in order for those methods to be able to discover, extract and represent significant and meaningful information \citep{Fernlund2006,Ngai2009}.

The largest body of research on EC in the KA consists of EC methods that can be included in the standard genetic algorithms (GA) category, either in single or multi-objective variants. The algorithms follow in general the broad guidelines of classic GAs, i.e. they use classic encoding methods such as binary or decimal (real, integer) with one feature in one chromosome, fixed sized populations with no sub-populations, genetic operators such as single-point and multi-point crossover and mutation, classic selection techniques such as roulette wheel selection or stochastic universal selection, etc. \citep{Goldberg1989}. Applications and reviews of these algorithms can be found in  \citep{Kuo2005}, \citep{Bandyopadhyay2007}, \citep{Dehuri2008}, \citep{Kim2010} and \citep{Mukhopadhyay2011}.

Apart from those considered standard GAs, a number of non-standard evolutionary techniques have been largely employed in KA tasks, such as the Non-dominated Sorting Genetic Algorithm \citep{Ishibuchi2005,Alcala2011}, the Niched Pareto Genetic Algorithm \citep{Emmanouilidis2009}, the Evolutionary Multi-Objective algorithm for Feature Selection \citep{Vatolkin2015}, or the Evolutionary Local Search Algorithm \citep{Kim2010,Zhou2011}.

A very recent and comprehensive review of the EC methods employed in KA is offered in \citep{Mukhopadhyay2014} and \citep{Mukhopadhyay2014a}, where the authors identify the most used evolutionary techniques for feature selection, classification, clustering and association rule mining, and provide detailed guidance for adapting the design of each component of the evolutionary algorithms (e.g. encoding, genetic operators, selection strategies, objective functions) to the desired knowledge acquisition task.

Similar to the other agent-enabling technologies, in the EC technology case too the machine agents can employ one or more EC techniques, which in turn can make use of either or both statistical methods and machine learning methods, resulting in complex integrated machine agents covering more than one stage/sub-process in the KA process.

\subsection{The mining machine agents}

The machine agents have been extensively used in KA as part of the autonomous KD paradigm. In \citep{Cao2007,Cao2009} the authors note that agents - generally seen through the lens of autonomous agents and multi-agent systems, and knowledge discovery - generally seen as a data mining exercise, initially emerged and established as separate standalone research fields but in the last two decades methods from both fields merged into a new field of research, the ``agent mining''. The agent mining concept was largely supported by numerous studies in the literature \citep{Balter2002,Cao2012,Moradi2013,Kadhim2014,ChemChem2015}, which proposed agents for agent mining applications under various names such as knowledge driven agents \citep{Balter2002}, knowledge collector agents \citep{Moradi2013}, or miner agents \citep{Albashiri2009,ChemChem2015}. In \citep{Cao2009} Cao et al. identify three approaches on agents and knowledge discovery that are essential for the emergence and establishment of the machine agents in knowledge acquisition: the data mining-driven agents, the agent-driven data mining, and the agent mining itself, where the former two were precursors of the latter. In the data mining-driven agents, the agents are empowered and their capabilities are enhanced by data mining and knowledge discovery techniques. Conversely, in the agent-driven data mining the agents or agent technologies in general are used to improve data mining processes. The agent mining encompasses both paradigms, and merges agent and multi-agent technologies with data mining and knowledge discovery techniques. This brief historical view on agent mining is important for understanding the convergence of the research interest towards autonomous knowledge discovery through the use of machine agents.

\begin{figure}[h]
\begin{center}
    \includegraphics[width=0.99\linewidth]{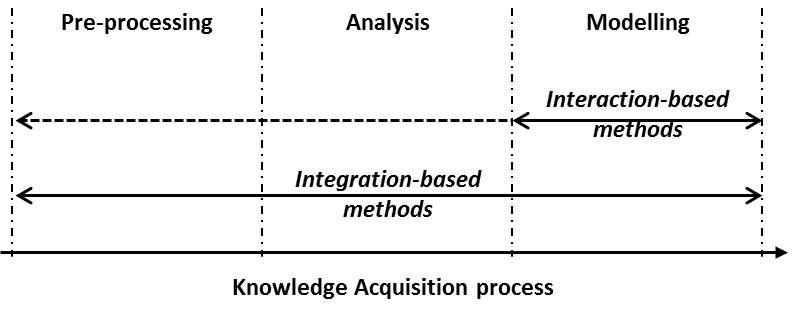}
    \caption{Mapping of he machine agent methods to the KA core process: solid line - ample explicit coverage, dashed line - limited implied coverage.}
    \label{Fig:MachineToProcess}
\end{center}
\end{figure}

However, from the point of view of this paper we are interested in the mechanics of the methods and therefore we identified in the literature two categories of machine agents for autonomous KD, the interaction-based and the integration-based agents, which we describe in the following. However, in order to guide the review of the literature on machine agency in KA, in Figure~\ref{Fig:MachineToProcess} we provide a mapping of the machine agent methods onto the KA process.

\subsubsection{Interaction-based agents}

The interaction-based methods present multi-agent approaches in which the learning mechanisms underlying the knowledge acquisition process are implemented using various types of social interactions and behaviours, such as collaboration, cooperation, negotiation, competition or imitation \citep{Guyet2007,Rybakov2009,Zhou2010,Ralha2012}. However, we found that of the many types of agent interactions reported in the general agent literature, only collaboration, cooperation and negotiation have been used in a significant amount for knowledge acquisition as part of the autonomous knowledge discovery paradigm, while the rest are only implied by some methods, rather than explicitly used as standalone methods.

In \citep{Guyet2007} the authors proposed a Collaborative Exploration System for knowledge discovery in biomedical multivariate time series data, in the form of a multi-agent system in which the subsequent learning is the result of cooperation between five types of agents (a data segmentation agent, a data classification agent, a symbolic translation agent, a scenario construction agent, and a system agent), which are grouped in two ``triads'': the symbolic triad and the system triad. In the symbolic triad the segmentation, classification and symbolic translation agents mutually interact to generate a symbolic representation of the raw data. In the system triad the symbolic interpretation, scenario construction and system agents interact to generate the knowledge representation. Further, a triad-to-triad interaction is also implemented in the form of a feedback cycle in order to refine and improve the resultant knowledge representation. Another collaborative system, MAS-KS (the Multi-Agent System based on Knowledge Sharing), was proposed by Schroeder and Bazzan \citep{Schroeder2002}, for improving individual learning models through knowledge sharing. In their approach each agent is derived from a machine learning algorithm which generates a set of rules. The method considers collaboration through a pair-wise interaction, in which two agents can match or merge their rules based on their performance in the knowledge extraction. Through this they update parameters of their models in order to improve the quality of the generated rules. A collaborative multi-agent system based on fitness-proportional knowledge sharing was also used in \citep{Leu2014} for cognitive skill modelling in the context of puzzle solving, where for the extraction of clue patterns from puzzles, procedural visual scanning skills must be developed by human players. The authors model computationally the cognitive skill acquisition process through a society of agents that search the puzzle grids and learn socially from the knowledge discovered and shared by fellow searching agents. An agent in the society updates its features by borrowing new features from another agent in the society considered to be its best fit. The amount of features borrowed is proportional to the fitness level, where the fitness considers agent feature compatibility and searching proficiency in the same time. In web search and web content mining, Chau et al. \citep{Chau2003} propose CS (the Collaborative Spider), a multi-agent system that allows user search collaboration for more accurate search results based on own and shared search history. Another collaborative environment was used in \citep{Ralha2012} for gaining knowledge about the dynamic strategies of companies involved in cartel formation. The authors introduced the AGent-MIning tool consisting of a society of collaborating agents grouped in specialised teams, and defined the interaction within and between the teams based on a standard released by the Foundation for Intelligent Physical Agents, called the ``FIPA Contract Net Interaction Protocol'' \citep{Agents2002}.

Cooperation and negotiation were also addressed in a number of studies \citep{Santos2005,Cheng2006,Lau2008}. A cooperative approach is used in \citep{Albashiri2009}, where Albashiri and colleagues introduce EMADS, an extendible multi-agent data mining system. The authors describe the proposed system as ``an anarchic collection of persistent, autonomous (but cooperating) KD agents operating across the Internet'', in which individual agents of different types, such as data agents, user agents, task agents, mining agents and ``house-keeping'' agents interact to provide pertinent knowledge models to whomever lodges a knowledge discovery request. In \citep{Santos2005} the authors use cooperative negotiation in a multi-agent distributed learning system consisting of several learning agents and a mediator agent, where the former implement different machine learning algorithms and the latter controls the negotiation-based interaction of the former. Using their own predefined subsets of the total available data, the learning agents extract rules and construct their own models of knowledge, which are then evaluated based on rule accuracy and used in negotiation. In the negotiation process the mediator agent collects proposals (models) from each learning agent, and filters them based on a threshold value. The models above the threshold are returned to the society of agents to be used in further search of rules with better accuracy. In \citep{Lau2008} a multi-agent system using adaptive probabilistic negotiation agents is proposed for knowledge discovery in e-marketplaces. The method uses a Bayesian learning mechanism that enables the agents to dynamically improve their negotiation skills by discovering key negotiation knowledge through mining and monitoring the interaction with and between the opponent agents. Further, a similar approach to negotiating agents has been presented in a different study by Cheng et al. \citep{Cheng2006} who discussed the concept of automated negotiation by autonomous agents. The automated negotiation agents jointly search for a mutually acceptable solution in the space formed by negotiable issues, where the negotiable issues are knowledge discovery aspects such as the fuzzy rules extracted by individual agents, if agents are seen as fuzzy inference systems.

\subsubsection{Integration-based agents}

The integration-based methods describe complex intelligent (autonomous) agents which fulfil the mining role individually, by implementing the learning mechanism implied by the knowledge discovery exercise through integration of one or more techniques from various fields, such as machine learning or evolutionary computation in order to implement one or more of the knowledge discovery tasks, such as rule extraction, classification or clustering \citep{Chen2001,Moradi2013,Kadhim2014}. From the integration point of view, Chemchem and Drias identify three major types of agents \citep{ChemChem2015}: agents based on expert systems, which use inference engines for constructing knowledge, agents based on machine learning, which extract knowledge using machine learning techniques, and agents based on data mining which rely on knowledge discovery methods for extracting the knowledge more efficiently. Most of the work on integrated mining agents uses various frameworks that in essence integrate in more or less extent these three major types of agents under different names, such as sub-agents, modules, components, units or processes \citep{Balter2002,Kadhim2014,ChemChem2015}.

In \citep{ChemChem2015}, based on the classification they proposed, the authors integrate the three approaches and introduce the Miner Intelligent Agent (MIA), a scalable knowledge-base cognitive agent that consists of a knowledge base, a meta-knowledge base, an induction rule mining module, and an inference engine. The knowledge base contains all the knowledge perceived and developed by the agent in the form of if-then rules. the rules are clustered and relations between the rules are found in the form of meta-rules. The meta-knowledge base contains the clustering information related to the clusters of knowledge, as well as the meta-rules associated to the clusters. The induction rule mining module continuously re-clusters new knowledge and the subsequent meta-knowledge, using two different K-means based algorithms for each of the knowledge base and meta-knowledge base. The Inference Engine creates the feedback loop that reinforce the rule base in order to emulate the reasoning process. Following the same type of integration framework as the one described in \citep{ChemChem2015}, Kadhim et al. \citep{Kadhim2014} proposed MIAKDD, a multi-intelligent agent for knowledge discovery in databases in cooperation with human experts. The MIAKDD agent integrates a combined rule generation-classification component which uses classification based on association. The output of rule generation and classification process is stored in a knowledge-base component which is accessed and reviewed by domain experts. The modified rules update the knowledge base, which is further accessed by the MIAKDD agent for refining the results.

In \citep{Fernlund2006}, the authors proposed an agent that integrates a genetic programming method for acquiring the knowledge underpinning tactical behaviours of both own and opponent forces in military and game simulations. The objective of the study was to learn the tactical behaviours of observed humans and create a tactical agent able to emulate those behaviours in a plausible manner. The genetic programming method was used in conjunction with context-based reasoning for evolving tactical agents using data acquired from humans performing missions on simulators. The agent proposed by Fernlund and colleagues was able to adopt the behaviour of the observed entity only from interpretation of the data collected through observation. A similar approach was used in \citep{Balter2002} where the authors proposed an integrated framework for knowledge discovery from electromyography (EMG) data. They used artefactual data from 1000 medical cases and over 25000 neurological tests for extracting pertinent medical information for diagnosis. Their miner agent framework uses a the EMG data module that ensures storage and update of raw data, a data-mining module that uses classic threshold-based association rule engine to gain knowledge about the medical domain, a knowledge-base module that stores the discovered knowledge, and a dissemination module that further refines the results in order to provide accurate individually customised feedback to users.

Fuzzy systems were also used in integrated mining agents. In \citep{Moradi2013} the authors propose a framework for capturing, storing, disseminating and utilising marketing knowledge, in which agent technology is combined with a fuzzy Analytical Hierarchy Process (fAHP) and fuzzy logic. The resultant fAHP miner agent uses the fuzzy AHP for allocating the weight of determinant criteria for the fuzzy rules, and the fuzzy logic refines the final decision for three situations: pessimistic, moderate and optimistic.

In \citep{Chen2001} the authors integrate a Bayesian network in a miner agent capable to discover causal relations between data objects in heterogeneous data sets. The proposed integrated agent environment embeds a data warehousing module, an on-line analytical processing (OLAP) module, and a knowledge discovery module for supporting the knowledge modelling tasks. The warehousing and OLAP modules implement the gathering, organising and storing information from various data resources. The knowledge discovery module uses a cross-reference technique for performing an informed search, and a Bayesian network for representing the association rules and knowledge patterns extracted from data. The authors note that the Bayesian network representation facilitates a non-monotonic reasoning process which, through integration with the other modules, facilitates data analysis (1) in a multidimensional data space and (2) at different levels of abstractions. A Bayesian approach was also used in \citep{Secretan2010}, where Secretan et al. propose an agent architecture for private and high-performance integrated data mining. The authors describe a miner agent that uses a Naive Bayes classifier for data classification and a data perturbation method - SMC (the secure multi-party computation), for privacy preserving in order to implement the knowledge discovery module, which in their study is called the privacy preserving data mining module.

\subsection{Summary of machine agents}

We included in this section those methods which suited the perspective depicted earlier in the paper in Figure~\ref{Fig:KAschema}. In Table~\ref{Table:summaryHA} we summarise this section in a tabular manner and through this, we emphasise the contribution of this class of methods to the creation of the roadmap that facilitates the KA process, and we show how each of the methods can fill an appropriate branch in the decision tree, corresponding to machine agency. Thus, overall, this section contributes to demonstrating the decision support role of the machine agent class of methods in the KA exercises.

\begin{table*}[!t]
	\centering
%	\begin{scriptsize}
		\begin{tabulary}{\linewidth}{|c|c|c|c|c|c|c|}
			\hline
			& \multicolumn{2}{c|}{\textbf{Agent perspective}} & \multicolumn{3}{c|}{\textbf{KA}} & \\
			\textbf{KA} & \multicolumn{2}{c|}{\textbf{roadmap}} & \multicolumn{3}{c|}{\textbf{process model}} & \textbf{References} \\ \cline{2-6} %\hline
			
			\textbf{approach} & \multicolumn{1}{c|}{\textbf{Amount of}} & \multicolumn{1}{c|}{\textbf{Richness of}} & \multicolumn{1}{c|}{\textbf{Elicitation}} & \multicolumn{1}{c|}{\textbf{Analysis}} & \multicolumn{1}{c|}{\textbf{Representation}} & \\ 
			
			& \multicolumn{1}{c|}{\textbf{activity}} & \multicolumn{1}{c|}{\textbf{analysis}} & \multicolumn{1}{c|}{} & \multicolumn{1}{c|}{} & \multicolumn{1}{c|}{} & \\ \hline

			Interaction-based & very high & rich & partial & partial & yes & \citep{Schroeder2002,Chau2003,Guyet2007,Ralha2012,Leu2014} \\
			Collaborative agents & & & & & & \\ \hline
			Interaction-based & very high & rich & partial & partial & yes & \citep{Santos2005,Albashiri2009}\\
			Cooperative agents & & & & & & \\ \hline
			Interaction-based & very high & rich & partial & partial & yes & \citep{Cheng2006,Lau2008} \\
			Negotiative agents & & & & & & \\ \hline
			Integration-based  & very high & very rich & yes & yes & yes & \citep{Balter2002,Fernlund2006,Kadhim2014,ChemChem2015} \\
			Miner Intelligent Agents & & & & & & \\ \hline
			Integration-based & very high & very rich & yes & yes & yes & \citep{Moradi2013} \\
			Fuzzy Miner Agents & & & & & & \\ \hline
			Integration-based& very high & very rich & yes & yes & yes & \citep{Chen2001,Secretan2010}\\
			Bayesian Miner Agents & & & & & & \\ \hline
		\end{tabulary}
%	\end{scriptsize}
	\caption{Summary of machine agent methods: agent perspective roadmap and KA process}
	\label{Table:summaryMA}
\end{table*}

\section{Discussion}

\subsection{The proposed perspective on KA}

The review provided in this paper captures the existing work on KA in human activity systems from an agent perspective, motivated by the continuous evolution of human activity over time. The section where we presented the historical view on KA and human activity showed how this evolution changed significantly the amount and the type of activity on one side, and the amount and availability of the SMEs and/or data that were subject to KA exercises. In these conditions, the KA exercise becomes subject to choosing not only the techniques, but in the first place the type of elicitor that will make use of those techniques. Thus, it becomes useful and pertinent to discuss the literature taking into consideration how the eliciting agency will be able to cope with the amount of activity to be analysed, given a certain level of analysis detail required by the goals of the KA exercise. We believe that having in place a classification that appropriately places the existing KA techniques into the portfolio of one or more of the human, human-inspired or machine agents, will allow one to decide the amount and the type of eliciting entities to be employed by the KA exercise, depending on the availability and size of the source of knowledge.

\begin{figure}[h]
	\begin{center}
		\includegraphics[width=0.99\linewidth]{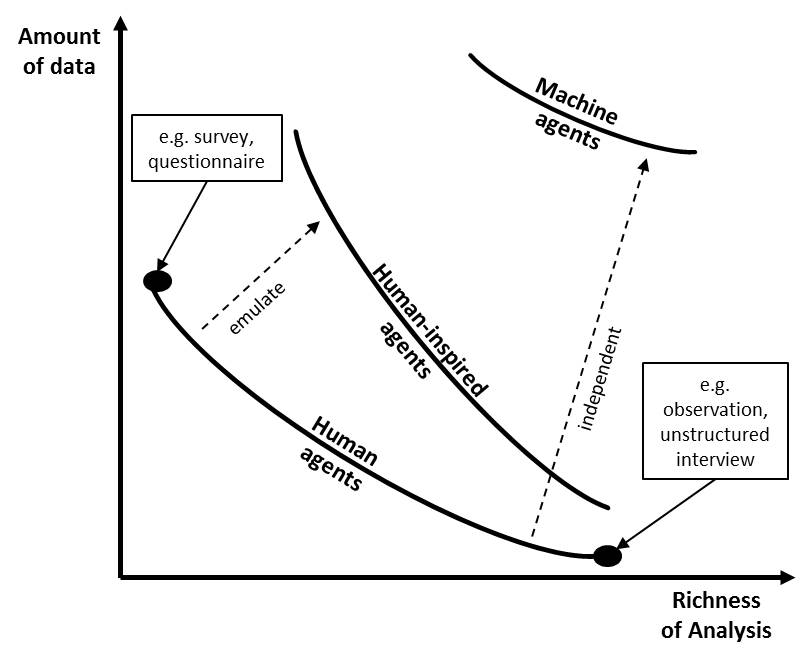}
		\caption{The proposed agent perspective: a data size - analysis/representation trade-off.}
		\label{Fig:Discussion1}
	\end{center}
\end{figure}

Early in the paper, before proceeding with the core review of the techniques for each agency, we provided in Figure~\ref{Fig:KAschema} a summary of this assumption and discussed how the proposed classification can be used as decision support for the KA exercise by facilitating the generation of a roadmap, in the form of a decision tree, associated to the KA process model. After reviewing the relevant techniques for each category of agents and summarising their role and scope in Table~\ref{Table:summaryHA}, Table~\ref{Table:summaryHIA} and Table~\ref{Table:summaryMA}, we are able to refine the assumption presented in Figure~\ref{Fig:KAschema}. Arguably, apart from providing decision support in the KA exercise, the proposed classification can describe the performances that can be achieved by each type of agent, taking into account the trade-off between the size of the knowledge source and the level of detail for analysis and representation. Figure~\ref{Fig:Discussion1} summarises this idea visually, depicting a trade-off curve for each agency.

The human agency has the lowest performance curve, due to obvious limitations in both analysis detail and knowledge source size. At one extreme, techniques like observations or unstructured interviews allow human elicitors to analyse SMEs and represent knowledge at high level of detail, however the amount of activity (i.e. number of SMEs) is drastically limited. For example, a human elicitor cannot handle large numbers of experts, such as hundreds of thousands, due to time and effort limitations. At the other extreme techniques like questionnaires or surveys can handle large numbers of subjects, eliminating much of the face-to-face elicitation time and effort; however, their contents cannot be excessively long or detailed. Thus, only low levels of detail for analysis and resultant representation can be achieved. Human-inspired agents can obtain a better trade-off curve, but only on the amount of activity (data) direction. In the activity/data direction, the number of questionnaires that can be analysed by an automated computer program is virtually unlimited compared to human analyst case. On the level of analysis direction though, due to the fact the human-inspired agents are mainly computer-based automated versions of the human elicitors, the highest level of analysis or representation detail cannot exceed that of human agents (i.e automated interviews compared to human elicitor interviews). Machine agent curve is, arguably, the one providing highest performances due to the ability of human-independent methods to analyse large amounts of data at high levels of detail. We consider that in this case both extremes of the curve can exceed their human and human-inspired counterparts through their very nature, generating better KA outputs in situations where the other types of agents have severe limitations (either technical or cognitive).

The trade-off view on the proposed classification completes the position we have regarding agents and the KA in human activity systems. Apart from summarising the contribution of the proposed classification as a model of choice in support for the KA exercise, the trade-off view also includes implicitly the historical evolution of the field and the KA process model. It includes the historical evolution through that it shows how the increase in the amount of activity (and resultant data) and the increasing level of analysis details created the need of new techniques, which moved the focus from human agents to human-inspired and then further to machine agents. It includes the KA process through that we discuss the trade-off using the elements of the three stage process model: elicitation (amount of activity/data direction), and analysis and representation (richness of analysis direction). 

\subsection{Future potential perspectives on KA}

The literature presented in this study suggests that the process of migrating from the classic human agent CTA methods to the human-independent machine agents is not only still open, but is far from being ended and has not yet reached its peak. We expect a significant increase in the conversion and application of the classical human agent methods to the new challenges raised by the broad concept of ``human activity'', in the conditions of higher and higher dynamics of change in this respect. Thus, we speculate on two possible directions of research that are candidates for inclusion in future reviews on ``human activity''-related KA, provided our expectation of growth materialises: the co-evolutionary KA, and the challenge-based KA.

We note that the categories we considered in this study, and the subsequent methods, do not take into account the change in the knowledge acquiring entity as a result of the interaction with the knowledge sourcing entity. While these methods may describe acquirer entities that can handle dynamic knowledge sources, they do not include mechanisms for treating the change in their own level of knowledge, that is, the way their perception about the acquired knowledge changes throughout the acquisition process, with an impact on the way they generate the ontological construct for representing the knowledge. We consider that the acquisition process is essentially a co-evolution process in which both parties evolve and gradually improve their understanding of the domain through interaction. Few very recent studies that relate co-evolutionary learning to knowledge acquisition exist \citep{Lotem2012,Wang2013}, however, reviews and explicit classification of the work are not yet in place. We expect a growth of the amount of work in this direction, which will lead in the future to significant advances in the knowledge acquisition.

Another aspect that we note in relation to the KA field is that the methods usually describe the ``extraction'' of knowledge from a knowledge sourcing entity. While certain degrees of interaction are described by the methods, this interaction does not reach the level where the knowledge acquiring entity challenges the knowledge sourcing entity (the SME) towards knowledge improvement, in order to be able to acquire better knowledge as a result of SME's improvement. A scenario explaining this concept is an elicitor that seeks knowledge about a domain but the available SME is not knowledgeable enough to satisfy elicitors' needs. Thus, the elicitor must first challenge the SME towards learning about the domain, in order to eventually acquire knowledge of a better quality. This approach was recently described as part of the Computational Red Teaming (CRT) paradigm \citep{Alam2012,Petraki2014,Abbass2015}, however the amount of work explicitly positioned as contribution to the KA field is still low. We expect that the future will bring an increased interest in this direction and we foresee a growth in the amount of published work.

Overall, we consider that the two research directions mentioned above, coevolution-based KA and CRT-based KA, can add in the future to the proposed agent view on KA by generating a complementary agent-based classification which captures two aspects of KA less present in the KA literature so far: (1) the amount of autonomy involved in the interaction between knowledge sourcing and knowledge acquiring entities and (2) the potential of the existing KA techniques to offer support for this interaction.

The first aspect refers to the ability of the agent to actively and autonomously seek for and fill the agreed ontological construct with the necessary and sufficient component elements. At one extreme passive agents can only receive the data that are made available to them, and thus can only fill the agreed ontological construct using those data. An example can be an automated interview in which a computer-based emulator of a human elicitor performs an interview with a SME. The emulator follows a fixed sequence of questions and receives answers from the SME, and is incapable of seeking clarifications if SME's answers do not fit in the agreed ontological construct. As a result, the knowledge representation will contain gaps corresponding to those parts of the ontological construct that were not filled due to incomplete or incorrect answers. At the other extreme, an active agent can seek autonomously for data that contribute to filling the gaps in the ontological construct, such as in the case of a human agent that performs an interview with an SME. In this case the human elicitor can actively seek for clarifications whenever SME's answers are not filling the ontological construct as required.

The second aspect refers to the potential offered by KA techniques to negotiate the ontological construct in order to achieve the agreed goals of KA exercise. An example can be acquisition of knowledge about diagnosing heart conditions in patients. If the only KA technique used for this purpose is a questionnaire using as an ontological construct the height and weight of the patient, then there is no chance of representing the knowledge about patients in a more detailed manner. However, if an unstructured interview is used for this purpose, then there is the potential of unveiling extra details due to the interaction between the elicitor and the patient, during which a set of elements that are essential for diagnosing the correct condition (i.e. the ontological construct) is gradually established. Thus, in this case the ontological construct is not fixed, being the result of a negotiation between the agents.

\begin{figure}[h]
	\begin{center}
		\includegraphics[width=0.99\linewidth]{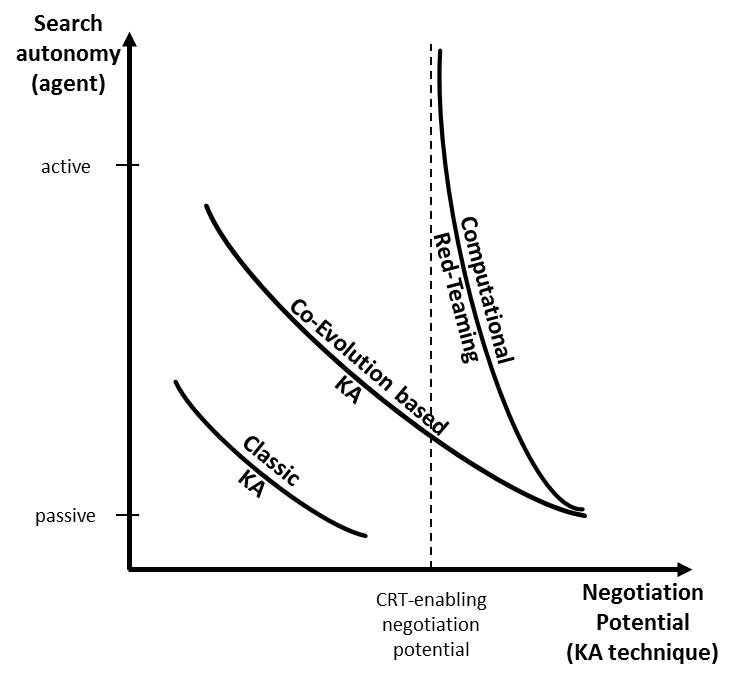}
		\caption{A future perspective on KA: the Autonomy-Negotiation trade-off.}
		\label{Fig:Discussion2}
	\end{center}
\end{figure}

We further consider that these two aspects are subject to a trade-off that generates a potential three-tier classification using co-evolution and Computational Red-Teaming. In Figure~\ref{Fig:Discussion2} we display this potential classification in the form of three trade-off curves: the lowest performance curve corresponds to what we could call the classic (or existent to date) views on KA, while the other two are the co-evolutionary and CRT-based views in increasing order of performance, respectively.

\section{Conclusions}

In this paper we provided a multi-disciplinary review which investigated from an agent perspective the knowledge acquisition in human activity systems, where knowledge acquisition is seen as the process of acquiring understanding about the underpinnings of human activity with the fundamental purpose of improving this activity. We proposed a classification of the methods based on the type of agency involved in the KA process, with a focus on the degree of involvement of human element. The motivation for this classification stays in the continuous change over time of the concept of human activity, as the factor that fuelled researchers' and practitioners' efforts for more than a century. Thus, we discussed the concept of knowledge acquisition under three categories - the human agents, the human-inspired agents, and the machine agents - and we showed how this classification can play a decision-support role in relation to the KA exercises.

% quick table code: http://www.tablesgenerator.com/
%\bibliographystyle{plainnat}
%\bibliography{CTAreviewBib}

\section*{Acknowledgement}

This work has been funded by the Australian Research Council (ARC) discovery grant, number DP140102590: Challenging systems to discover vulnerabilities using computational red teaming.

This is a pre-print of an article published in Knowledge-Based Systems, vol. 105, Elsevier. The final authenticated version is available online at: https://doi.org/10.1016/j.knosys.2016.02.012

\section*{References}

\bibliographystyle{elsarticle-harv}
%\bibliography{KBreviewBib}

\begin{thebibliography}{216}
	\expandafter\ifx\csname natexlab\endcsname\relax\def\natexlab#1{#1}\fi
	\expandafter\ifx\csname url\endcsname\relax
	\def\url#1{\texttt{#1}}\fi
	\expandafter\ifx\csname urlprefix\endcsname\relax\def\urlprefix{URL }\fi
	
	\bibitem[{Abbass(2015)}]{Abbass2015}
	Abbass, H.~A., 2015. Computational Red Teaming. Springer International
	Publishing.
	
	\bibitem[{Afifi and Azen(1972)}]{Afifi1972}
	Afifi, A.~A., Azen, S.~P., 1972. Statistical analysis: a computer oriented
	approach. Academic Press, NY.
	
	\bibitem[{Agents(2002)}]{Agents2002}
	Agents, F. f. I.~P., 03.12.2002 2002. Fipa contract net interaction protocol.
	
	\bibitem[{Ahn and Yager(2014)}]{Ahn2014}
	Ahn, B.~S., Yager, R.~R., 2014. The use of ordered weighted averaging method
	for decision making under uncertainty. International Transactions in
	Operational Research 21~(2), 247--262.
	
	\bibitem[{Alam et~al.(2012)Alam, Zhao, Tang, Lokan, Ellejmi, Kirby, and
		Abbass}]{Alam2012}
	Alam, S., Zhao, W., Tang, J., Lokan, C., Ellejmi, M., Kirby, S., Abbass, H.,
	2012. Discovering delay patterns in arrival traffic with dynamic continuous
	descent approaches using co-evolutionary red teaming. Air Traffic Control
	Quarterly 20~(1), 47.
	
	\bibitem[{Albashiri et~al.(2009)Albashiri, Coenen, and Leng}]{Albashiri2009}
	Albashiri, K.~A., Coenen, F., Leng, P., 2009. Emads: An extendible multi-agent
	data miner. Knowledge-Based Systems 22~(7), 523 -- 528, artificial
	Intelligence 2008 AI-2008 The twenty-eighth \{SGAI\} International Conference
	on Artificial Intelligence.
	
	\bibitem[{Alcala et~al.(2011)Alcala, Nojima, Herrera, and
		Ishibuchi}]{Alcala2011}
	Alcala, R., Nojima, Y., Herrera, F., Ishibuchi, H., 2011. Multiobjective
	genetic fuzzy rule selection of single granularity-based fuzzy classification
	rules and its interaction with the lateral tuning of membership functions.
	Soft Computing 15~(12), 2303--2318.
	
	\bibitem[{Anderson(1983)}]{Anderson1983}
	Anderson, J.~R., 1983. A spreading activation theory of memory. Journal of
	Verbal Learning and Verbal Behavior 22~(3), 261 -- 295.
	
	\bibitem[{Anderson(1996)}]{Anderson1996}
	Anderson, J.~R., 1996. Act: A simple theory of complex cognition. American
	Psychologist 51, 355--365.
	
	\bibitem[{Antonova and Stefanov(2011)}]{Antonova2011}
	Antonova, A., Stefanov, K., 2011. Applied cognitive task analysis in the
	context of serious games development. In: Dicheva, D., Markov, Z., Stefanova,
	E. (Eds.), Third International Conference on Software, Services and Semantic
	Technologies S3T 2011. Vol. 101 of Advances in Intelligent and Soft
	Computing. Springer Berlin Heidelberg, pp. 175--182.
	
	\bibitem[{Arend(1991)}]{Arend1991}
	Arend, U., 1991. Analyzing complex tasks with an extended GOMS model. Elsevier,
	B.V. North-Holland, pp. 115--133.
	
	\bibitem[{Aussenac-Gilles and Matta(1994)}]{Aussenac-Gilles1994}
	Aussenac-Gilles, N., Matta, N., 1994. Making a method of problem solving
	explicit with \{MACAO\}. International Journal of Human-Computer Studies
	40~(2), 193 -- 219.
	
	\bibitem[{Bainbridge(1979)}]{Bainbridge1979}
	Bainbridge, L., 1979. Verbal reports as evidence of the process operator's
	knowledge. International Journal of Man-Machine Studies 11, 411--436.
	
	\bibitem[{Bainbridge(2007)}]{Bainbridge2007}
	Bainbridge, W.~S., 2007. The scientific research potential of virtual worlds.
	Science 317~(5837), 472--476.
	
	\bibitem[{Balter et~al.(2002)Balter, Labarre-Vila, Ziebelin, and
		Garbay}]{Balter2002}
	Balter, J., Labarre-Vila, A., Ziebelin, D., Garbay, C., 2002. A
	knowledge-driven agent-centred framework for data mining in \{EMG\}. Comptes
	Rendus Biologies 325~(4), 375 -- 382.
	
	\bibitem[{Bandyopadhyay et~al.(2007)Bandyopadhyay, Maulik, and
		Mukhopadhyay}]{Bandyopadhyay2007}
	Bandyopadhyay, S., Maulik, U., Mukhopadhyay, A., May 2007. Multiobjective
	genetic clustering for pixel classification in remote sensing imagery.
	Geoscience and Remote Sensing, IEEE Transactions on 45~(5), 1506--1511.
	
	\bibitem[{Beach and Pedersen(2013)}]{Beach2013}
	Beach, D., Pedersen, R.~B., 2013. Process-Tracing Methods: Foundations and
	Guidelines. University of Michigan Press, Ann Arbor:.
	
	\bibitem[{Begoli and Horey(2012)}]{Begoli2012}
	Begoli, E., Horey, J., Aug 2012. Design principles for effective knowledge
	discovery from big data. In: Software Architecture (WICSA) and European
	Conference on Software Architecture (ECSA), 2012 Joint Working IEEE/IFIP
	Conference on. pp. 215--218.
	
	\bibitem[{Benjamini and Leshno(2010)}]{Benjamini2010}
	Benjamini, Y., Leshno, M., 2010. Statistical methods for data mining. In:
	Maimon, O., Rokach, L. (Eds.), Data Mining and Knowledge Discovery Handbook.
	Springer US, pp. 523--540.
	
	\bibitem[{Beringer and Wandmacher(1991)}]{Beringer1991}
	Beringer, J., Wandmacher, J., 1991. Object-based action planning. Vol.~2.
	Elsevier, B.V. North-Holland, pp. 135--155.
	
	\bibitem[{Bogo et~al.(2014)Bogo, Shlonsky, Lee, and Serbinski}]{Bogo2014}
	Bogo, M., Shlonsky, A., Lee, B., Serbinski, S., 2014. Acting like it matters: A
	scoping review of simulation in child welfare training. Journal of Public
	Child Welfare 8~(1), 70--93.
	
	\bibitem[{Boose(1989)}]{Boose1989a}
	Boose, J.~H., 1989. A survey of knowledge acquisition techniques and tools.
	Knowledge Acquisition 1~(1), 3 -- 37.
	
	\bibitem[{Boose et~al.(1989)Boose, Shema, and Bradshaw}]{Boose1989}
	Boose, J.~H., Shema, D.~B., Bradshaw, J.~M., 1989. Recent progress in aquinas:
	A knowledge acquisition workbench. Knowledge Acquisition 1~(2), 185 -- 214.
	
	\bibitem[{Borg and Groenen(2005)}]{Borg2005}
	Borg, I., Groenen, P. J.~F., 2005. Modern multidimensional scaling: Theory and
	applications, 2nd Edition. Springer Series in Statistics. Springer-Verlag New
	York.
	
	\bibitem[{Bradshaw and Boose(1990)}]{Bradshaw1990}
	Bradshaw, J.~M., Boose, J.~H., 1990. Decision analysis techniques for knowledge
	acquisition: Combining information and preferences using aquinas and axotl.
	International Journal of Man-Machine Studies 32~(2), 121--186.
	
	\bibitem[{Bradshaw et~al.(1993)Bradshaw, Ford, Adams-Webber, and
		Boose}]{Bradshaw1993}
	Bradshaw, J.~M., Ford, K.~M., Adams-Webber, J.~R., Boose, J.~H., 1993. Beyond
	the repertory grid: new approaches to constructivist knowledge acquisition
	tool development. International Journal of Intelligent Systems 8~(2),
	287--333.
	
	\bibitem[{Brubacher et~al.(2015)Brubacher, Powell, Skouteris, and
		Guadagno}]{Brubacher2015}
	Brubacher, S.~P., Powell, M., Skouteris, H., Guadagno, B., 2015. The effects of
	e-simulation interview training on teachers' use of open-ended questions.
	Child Abuse \& Neglect accepted - in press~(0), --.
	
	\bibitem[{Byrne and Anderson(1997)}]{Byrne1997}
	Byrne, M.~D., Anderson, J.~R., 1997. Enhancing act-r's perceptual-motor
	abilites. In: the 19th Annual Conference of the Cognitive Science Society.
	Mahwah, NJ: Erlbaum, p. 880.
	
	\bibitem[{Cano et~al.(2003)Cano, Herrera, and Lozano}]{Cano2003}
	Cano, J., Herrera, F., Lozano, M., Dec 2003. Using evolutionary algorithms as
	instance selection for data reduction in kdd: an experimental study.
	Evolutionary Computation, IEEE Transactions on 7~(6), 561--575.
	
	\bibitem[{Cao et~al.(2009)Cao, Gorodetsky, and Mitkas}]{Cao2009}
	Cao, L., Gorodetsky, V., Mitkas, P., May 2009. Agent mining: The synergy of
	agents and data mining. Intelligent Systems, IEEE 24~(3), 64--72.
	
	\bibitem[{Cao et~al.(2007)Cao, Luo, and Zhang}]{Cao2007}
	Cao, L., Luo, C., Zhang, C., 2007. Agent-mining interaction: An emerging area.
	In: Gorodetsky, V., Zhang, C., Skormin, V., Cao, L. (Eds.), Autonomous
	Intelligent Systems: Multi-Agents and Data Mining. Vol. 4476 of Lecture Notes
	in Computer Science. Springer Berlin Heidelberg, pp. 60--73.
	
	\bibitem[{Cao et~al.(2012)Cao, Weiss, and Yu}]{Cao2012}
	Cao, L., Weiss, G., Yu, P., 2012. A brief introduction to agent mining.
	Autonomous Agents and Multi-Agent Systems 25~(3), 419--424.
	
	\bibitem[{Card et~al.(1983)Card, Moran, and Newell}]{Card1983}
	Card, S.~K., Moran, T.~P., Newell, A.~L., 1983. The psychology of the
	human-computer interface. Lawrence Erlbaum Associates, Hillsdale, NJ.
	
	\bibitem[{Card et~al.(1986)Card, Moran, and Newell}]{Card1986}
	Card, S.~K., Moran, T.~P., Newell, A.~L., 1986. The model human processor: an
	engineering model of human performance. Vol.~2. John Wiley and Sons, New
	York.
	
	\bibitem[{Carpenter et~al.(2005)Carpenter, Martens, and Ogas}]{Carpenter2005}
	Carpenter, G.~A., Martens, S., Ogas, O.~J., 2005. Self-organizing information
	fusion and hierarchical knowledge discovery: a new framework using artmap
	neural networks. Neural Networks 18~(3), 287--295.
	
	\bibitem[{Carroll and Olson(1988)}]{Carroll1988}
	Carroll, J., Olson, J.~R., 1988. Handbook of human-computer interaction.
	Amsterdam: Elsevier, Ch. Mental models in human-computer interaction.
	
	\bibitem[{Chao et~al.(1999)Chao, Salvendy, and Lightner}]{Chao1999}
	Chao, C., Salvendy, G., Lightner, N.~J., 1999. Development of a methodology for
	optimizing elicited knowledge. Behaviour \& Information Technology 18~(6),
	413--430.
	
	\bibitem[{Chattopadhyay et~al.(2014)Chattopadhyay, Dan, and
		Mazumdar}]{Chattopadhyay2014}
	Chattopadhyay, M., Dan, P.~K., Mazumdar, S., 2014. Comparison of visualization
	of optimal clustering using self-organizing map and growing hierarchical
	self-organizing map in cellular manufacturing system. Applied Soft Computing
	22~(0), 528 -- 543.
	
	\bibitem[{Chau et~al.(2003)Chau, Zeng, Chen, Huang, and Hendriawan}]{Chau2003}
	Chau, M., Zeng, D., Chen, H., Huang, M., Hendriawan, D., 2003. Design and
	evaluation of a multi-agent collaborative web mining system. Decision Support
	Systems 35~(1), 167 -- 183, web Retrieval and Mining.
	
	\bibitem[{Chemchem and Drias(2015)}]{ChemChem2015}
	Chemchem, A., Drias, H., 2015. From data mining to knowledge mining:
	Application to intelligent agents. Expert Systems with Applications 42~(3),
	1436 -- 1445.
	
	\bibitem[{Chen and Pu(2012)}]{Chen2012}
	Chen, L., Pu, P., 2012. Critiquing-based recommenders: survey and emerging
	trends. User Modeling and User-Adapted Interaction 22~(1-2), 125--150.
	
	\bibitem[{Chen et~al.(2001)Chen, Zhu, and Chen}]{Chen2001}
	Chen, M., Zhu, Q., Chen, Z., 2001. An integrated interactive environment for
	knowledge discovery from heterogeneous data resources. Information and
	Software Technology 43~(8), 487 -- 496.
	
	\bibitem[{Cheng et~al.(2006)Cheng, Chan, and Lin}]{Cheng2006}
	Cheng, C.-B., Chan, C.-C.~H., Lin, K.-C., 2006. Intelligent agents for
	e-marketplace: Negotiation with issue trade-offs by fuzzy inference systems.
	Decision Support Systems 42~(2), 626 -- 638.
	
	\bibitem[{Cheung et~al.(2007)Cheung, Li, Shek, Lee, and Tsang}]{Cheung2007}
	Cheung, C., Li, M., Shek, W., Lee, W., Tsang, T., 2007. A systematic approach
	for knowledge auditing: a case study in transportation sector. Journal of
	Knowledge Management 11~(4), 140--158.
	
	\bibitem[{Clark(2014)}]{Clark2014}
	Clark, R., 2014. Cognitive task analysis for expert-based instruction in
	healthcare. In: Spector, J.~M., Merrill, M.~D., Elen, J., Bishop, M.~J.
	(Eds.), Handbook of Research on Educational Communications and Technology.
	Springer New York, pp. 541--551.
	
	\bibitem[{Clark et~al.(2008)Clark, Feldon, Van~Merrienboer, Yates, and
		Early}]{Clark2008}
	Clark, R.~E., Feldon, D., Van~Merrienboer, J. J.~G., Yates, K., Early, S.,
	2008. Cognitive task analysis, 3rd Edition. Mahwah, NJ: Lawrence Erlbaum
	Associates.
	
	\bibitem[{Code et~al.(2012)Code, Clarke-Midura, Zap, and Dede}]{Code2012}
	Code, J., Clarke-Midura, J., Zap, N., Dede, C., 2012. Virtual performance
	assessment in immersive virtual environments. Information Science Reference
	(IGI Global), Hershey, PA, pp. 230--252.
	
	\bibitem[{CogAff(2013)}]{CogAff2013}
	CogAff, 2013.
	\newline\urlprefix\url{http://www.cs.bham.ac.uk/research/projects/cogaff/}
	
	\bibitem[{Cooke(1994)}]{Cooke1994}
	Cooke, N.~J., 1994. Varieties of knowledge elicitation techniques.
	International Journal of Human-Computer Studies 41~(6), 801â€“849.
	
	\bibitem[{Corapi et~al.(2012)Corapi, Russo, and Lupu}]{Corapi2012}
	Corapi, D., Russo, A., Lupu, E., 2012. Inductive logic programming in answer
	set programming. In: Muggleton, S.~H., Tamaddoni-Nezhad, A., Lisi, F.~A.
	(Eds.), Inductive Logic Programming. Vol. 7207 of Lecture Notes in Computer
	Science. Springer Berlin Heidelberg, pp. 91--97.
	
	\bibitem[{Crandall et~al.(2006)Crandall, Klein, and Hoffman}]{Crandall2006}
	Crandall, B., Klein, G., Hoffman, R.~R., 2006. Working minds: A practitioner's
	guide to cognitive task analysis. MIT Press, Cambridge, MA.
	
	\bibitem[{Crilly et~al.(2006)Crilly, Blackwell, and Clarkson}]{Crilly2006}
	Crilly, N., Blackwell, A.~F., Clarkson, P.~J., 2006. Graphic elicitation: using
	research diagrams as interview stimuli. Qualitative Research 6~(3), 341--366.
	
	\bibitem[{Cullen and Bryman(1988)}]{Cullen1988}
	Cullen, J., Bryman, A., 1988. The knowledge acquisition bottleneck: Time for
	reassessment? Expert Systems 5~(3), 216--225.
	
	\bibitem[{Daghfous et~al.(2013)Daghfous, Ahmad, and Angell}]{Daghfous2013}
	Daghfous, A., Ahmad, N., Angell, L.~C., 2013. The kcrm knowledge audit: model
	and case illustration. VINE 43~(2), 185--209.
	
	\bibitem[{David and David(2014)}]{David2014}
	David, F., David, C.~M., 2014. Adapting cognitive walkthrough to support game
	based learning design. International Journal of Game-Based Learning (IJGBL)
	4~(3), 23--34.
	
	\bibitem[{Davidsson and Alm(2014)}]{Davidsson2014}
	Davidsson, S., Alm, H., 2014. Context adaptable driver information - or, what
	do whom need and want when? Applied Ergonomics 45~(4), 994 -- 1002.
	
	\bibitem[{Davis et~al.(2006)Davis, Dieste, Hickey, Juristo, and
		Moreno}]{Davis2006}
	Davis, A., Dieste, O., Hickey, A., Juristo, N., Moreno, A.~M., 2006.
	Effectiveness of requirements elicitation techniques: Empirical results
	derived from a systematic review. In: Requirements Engineering, 14th IEEE
	International Conference.
	
	\bibitem[{Dehuri et~al.(2008)Dehuri, Patnaik, Ghosh, and Mall}]{Dehuri2008}
	Dehuri, S., Patnaik, S., Ghosh, A., Mall, R., 2008. Application of elitist
	multi-objective genetic algorithm for classification rule generation. Applied
	Soft Computing 8~(1), 477 -- 487.
	
	\bibitem[{Diederich et~al.(1987)Diederich, Ruhmann, and May}]{Diederich1987}
	Diederich, J., Ruhmann, I., May, M., 1987. Kriton: a knowledge-acquisition tool
	for expert systems. International Journal of Man-Machine Studies 26~(1), 29
	-- 40.
	
	\bibitem[{dos Santos and Bazzan(2005)}]{Santos2005}
	dos Santos, C., Bazzan, A., 2005. Integrating knowledge through cooperative
	negotiation - a case study in bioinformatics. In: Gorodetsky, V., Liu, J.,
	Skormin, V. (Eds.), Autonomous Intelligent Systems: Agents and Data Mining.
	Vol. 3505 of Lecture Notes in Computer Science. Springer Berlin Heidelberg,
	pp. 277--288.
	
	\bibitem[{Draper and Smith(2014)}]{Draper2014}
	Draper, N.~R., Smith, H., 2014. Applied regression analysis, 3rd Edition. John
	Wiley \& Sons.
	
	\bibitem[{Driessnack and Furukawa(2012)}]{Driessnack2012}
	Driessnack, M., Furukawa, R., 2012. Arts-based data collection techniques used
	in child research. Journal for Specialists in Pediatric Nursing 17~(1), 3--9.
	
	\bibitem[{Drury(1990)}]{Drury1990}
	Drury, C.~G., 1990. Evaluation of human work: a practical ergonomics
	methodology. London: Taylor \& Francis, Ch. Methods for direct observation of
	performance, pp. 35--57.
	
	\bibitem[{Durbach and Stewart(2012)}]{Durbach2012}
	Durbach, I.~N., Stewart, T.~J., 2012. Modeling uncertainty in multi-criteria
	decision analysis. European Journal of Operational Research 223~(1), 1 -- 14.
	
	\bibitem[{Eberts(1997)}]{Eberts1997}
	Eberts, R., 1997. Cognitive Modeling, 2nd Edition. John Wiley \& Sons, New
	York, book section~40, pp. 1--47.
	
	\bibitem[{Elkerton and Palmiter(1991)}]{Elkerton1991}
	Elkerton, J., Palmiter, S.~L., 1991. Designing help using a goms model: An
	information retrieval evaluation. Human Factors: The Journal of the Human
	Factors and Ergonomics Society 33~(2), 185--204.
	
	\bibitem[{Embrey(2000)}]{Embrey2000}
	Embrey, D., 2000. Task analysis techniques. Tech. rep., Human Realiability
	Associates.
	
	\bibitem[{Emmanouilidis et~al.(2009)Emmanouilidis, Batsalas, and
		Papamarkos}]{Emmanouilidis2009}
	Emmanouilidis, C., Batsalas, C., Papamarkos, N., July 2009. Development and
	evaluation of text localization techniques based on structural texture
	features and neural classifiers. In: Document Analysis and Recognition, 2009.
	ICDAR '09. 10th International Conference on. pp. 1270--1274.
	
	\bibitem[{Eshelman et~al.(1987)Eshelman, Ehret, McDermott, and
		Tan}]{Eshelman1987}
	Eshelman, L., Ehret, D., McDermott, J., Tan, M., 1987. Mole: a tenacious
	knowledge-acquisition tool. International Journal of Man-Machine Studies
	26~(1), 41 -- 54.
	
	\bibitem[{Essens et~al.(1995)Essens, Fallesen, McCann, Cannon-Bowers, and
		Dorfel}]{Essens1995}
	Essens, P.~J., Fallesen, J.~J., McCann, C.~A., Cannon-Bowers, J.~A., Dorfel,
	G., 1995. Coade: A framework for cognitive analysis, design, and evaluation.
	Tech. rep., Brussels, Belgium: NATO Defence Research Group.
	
	\bibitem[{Fan et~al.(2012)Fan, Kalyanpur, Gondek, and Ferrucci}]{Fan2012}
	Fan, J., Kalyanpur, A., Gondek, D.~C., Ferrucci, D.~A., 2012. IBM J. Res. \&
	Dev. 56~(3/4).
	
	\bibitem[{Fernlund et~al.(2006)Fernlund, Gonzalez, Georgiopoulos, and
		DeMara}]{Fernlund2006}
	Fernlund, H. K. .~G., Gonzalez, A.~J., Georgiopoulos, M., DeMara, R.~F., Feb
	2006. Learning tactical human behavior through observation of human
	performance. Systems, Man, and Cybernetics, Part B: Cybernetics, IEEE
	Transactions on 36~(1), 128--140.
	
	\bibitem[{Fisk and Eggemeier(1988)}]{Fisk1988}
	Fisk, A.~D., Eggemeier, F.~T., 1988. Application of automatic/controlled
	processing theory to training tactical command and control skills: 1.
	background and task analytic methodology. Proceedings of the Human Factors
	and Ergonomics Society Annual Meeting 32~(18), 1227--1231.
	
	\bibitem[{Ford et~al.(1993)Ford, Bradshaw, Adams-Webber, and Agnew}]{Ford1993}
	Ford, K.~M., Bradshaw, J.~M., Adams-Webber, J.~R., Agnew, N.~M., 1993.
	Knowledge acquisition as a constructive modeling activity. International
	Journal of Intelligent Systems 8~(1), 9--32.
	
	\bibitem[{Fox et~al.(2011)Fox, Ericsson, and Best}]{Fox2011}
	Fox, M.~C., Ericsson, K.~A., Best, R., 2011. Do procedures for verbal reporting
	of thinking have to be reactive? a meta-analysis and recommendations for best
	reporting methods. Psychological Bulletin 137~(2), 316--344.
	
	\bibitem[{Gaines and Shaw(1992)}]{Gaines1992}
	Gaines, B., Shaw, M., 1992. Integrated knowledge acquisition architectures.
	Journal of Intelligent Information Systems 1~(1), 9--34.
	
	\bibitem[{Gallagher(1979)}]{Gallagher1979}
	Gallagher, J., 1979. Cognitive/information processing psychology and
	instruction: Reviewing recent theory and practice. Instructional Science
	8~(4), 393--414.
	
	\bibitem[{Gazarian(2013)}]{Gazarian2013}
	Gazarian, P.~K., 2013. Use of the critical decision method in nursing research:
	An integrative review. Advances in Nursing Science 36~(2), 106--117.
	
	\bibitem[{Geis et~al.(2013)Geis, Wheeler, Bunger, Taylor, Militello, and
		Patterson}]{Geis2013}
	Geis, G., Wheeler, D., Bunger, A., Taylor, R., Militello, L., Patterson, M.,
	2013. 137: Leveraging critical decision method and simulation-based training
	to accelerate sepsis recognition. Critical Care Medicine 41~(12), A28.
	
	\bibitem[{Geiwitz et~al.(1988)Geiwitz, Klatsky, and P.}]{Geiwitz1988}
	Geiwitz, J., Klatsky, R.~L., P., M.~B., 1988. Knowledge acquisition for expert
	systems: conceptual and empirical comparisons. Anacapa Sciences, Santa
	Barbara, CA.
	
	\bibitem[{Goldberg(1989)}]{Goldberg1989}
	Goldberg, D.~E., 1989. Genetic Algorithms in Search, Optimization and Machine
	Learning, 1st Edition. Addison-Wesley Longman Publishing Co., Inc., Boston,
	MA, USA.
	
	\bibitem[{Gordon et~al.(1993)Gordon, Schmierer, and Gill}]{Gordon1993}
	Gordon, S.~E., Schmierer, K.~A., Gill, R.~T., 1993. Conceptual graph analysis:
	knowledge acquisition for instructional systems design. Human Factors 35,
	459--481.
	
	\bibitem[{Gourova et~al.(2012)Gourova, Toteva, and Todorova}]{Gourova2012}
	Gourova, E., Toteva, K., Todorova, Y., 2012. Audit of knowledge flows and
	critical business processes. In: Proceedings of the 17th European Conference
	on Pattern Languages of Programs. EuroPLoP '12. ACM, New York, NY, USA, pp.
	1:1--1:10.
	
	\bibitem[{Govaerts et~al.(2013)Govaerts, Van~de Wiel, Schuwirth, Van~der
		Vleuten, and Muijtjens}]{Govaerts2013}
	Govaerts, M., Van~de Wiel, M., Schuwirth, L., Van~der Vleuten, C., Muijtjens,
	A., 2013. Workplace-based assessment: raters' performance theories and
	constructs. Advances in Health Sciences Education 18~(3), 375--396.
	
	\bibitem[{Gray(2008)}]{Gray2008}
	Gray, W.~D., 2008. Cognitive architectures: Choreographing the dance of mental
	operations with the task environment. Human Factors: The Journal of the Human
	Factors and Ergonomics Society 50~(3), 497--505.
	
	\bibitem[{Gray et~al.(1993)Gray, John, and Atwood}]{Gray1993}
	Gray, W.~D., John, B.~E., Atwood, M.~E., 1993. Project ernestine: Validating a
	goms analysis for predicting and explaining real-world task performance.
	Human-Computer Interaction 8~(3), 237--309.
	
	\bibitem[{Gruber(1989{\natexlab{a}})}]{Gruber1989a}
	Gruber, T.~R., 1989{\natexlab{a}}. The acquisition of strategic knowledge.
	Vol.~4 of Perspectives in Artificial Intelligence. Academic Press.
	
	\bibitem[{Gruber(1989{\natexlab{b}})}]{Gruber1989}
	Gruber, T.~R., 1989{\natexlab{b}}. Automated knowledge acquisition for
	strategic knowledge. Machine Learning 4~(3-4), 293--336.
	
	\bibitem[{Guhde(2014)}]{Guhde2014}
	Guhde, J.~A., 2014. An evaluation tool to measure interdisciplinary critical
	incident verbal reports. Nursing Education Perspectives 35~(3), 180--184.
	
	\bibitem[{Guyet et~al.(2007)Guyet, Garbay, and Dojat}]{Guyet2007}
	Guyet, T., Garbay, C., Dojat, M., 2007. Knowledge construction from time series
	data using a collaborative exploration system. Journal of Biomedical
	Informatics 40~(6), 672 -- 687, intelligent Data Analysis in Biomedicine.
	
	\bibitem[{Hall et~al.(1994)Hall, Gott, and Pokorny}]{Hall1994}
	Hall, E.~M., Gott, S.~P., Pokorny, R.~A., 1994. A procedural guide to cognitive
	task analysis: the PARI methodology. Brooks AFB, TX.
	
	\bibitem[{Hall et~al.(1995)Hall, Gott, and Pokorny}]{Hall1995}
	Hall, E.~P., Gott, S.~P., Pokorny, R.~A., 1995. A procedural guide to cognitive
	task analysis: The pari methodology. Report, Air Force Materiel Command.
	
	\bibitem[{Harzing et~al.(2009)Harzing, Baldueza, Barner-Rasmussen, Barzantny,
		Canabal, Davila, Espejo, Ferreira, Giroud, Koester, Liang, Mockaitis, Morley,
		Myloni, Odusanya, O'Sullivan, Palaniappan, Prochno, Choudhury, Saka-Helmhout,
		Siengthai, Viswat, Soydas, and Zander}]{Harzing2009}
	Harzing, A.-W., Baldueza, J., Barner-Rasmussen, W., Barzantny, C., Canabal, A.,
	Davila, A., Espejo, A., Ferreira, R., Giroud, A., Koester, K., Liang, Y.-K.,
	Mockaitis, A., Morley, M.~J., Myloni, B., Odusanya, J.~O., O'Sullivan, S.~L.,
	Palaniappan, A.~K., Prochno, P., Choudhury, S.~R., Saka-Helmhout, A.,
	Siengthai, S., Viswat, L., Soydas, A.~U., Zander, L., 2009. Rating versus
	ranking: What is the best way to reduce response and language bias in
	cross-national research? International Business Review 18~(4), 417 -- 432.
	
	\bibitem[{Hasler et~al.(2013)Hasler, Tuchman, and Friedman}]{Hasler2013}
	Hasler, B.~S., Tuchman, P., Friedman, D., 2013. Virtual research assistants:
	Replacing human interviewers by automated avatars in virtual worlds.
	Computers in Human Behavior 29~(4), 1608 -- 1616.
	
	\bibitem[{Hoffman(1992)}]{Hoffman1992}
	Hoffman, R., 1992. Doing psychology in an ai context: A personal perspective
	and introduction to this volume. In: Hoffman, R. (Ed.), The Psychology of
	Expertise. Springer New York, pp. 3--11.
	
	\bibitem[{Hoffman(1987)}]{Hoffman1987}
	Hoffman, R.~R., 1987. The problem of extracting the knowledge of experts from
	the perspective of experimental-psychology. Ai Magazine 8~(2), 53--67.
	
	\bibitem[{Hoffman et~al.(1998)Hoffman, Crandall, and Shadbolt}]{Hoffman1998}
	Hoffman, R.~R., Crandall, B., Shadbolt, N., 1998. Use of the critical decision
	method to elicit expert knowledge: A case study in the methodology of
	cognitive task analysis. Human Factors: The Journal of the Human Factors and
	Ergonomics Society 40~(2), 254--276.
	
	\bibitem[{Hoffman and Militello(2008)}]{Hoffman2008}
	Hoffman, R.~R., Militello, L., 2008. Perspectives on cognitive task analysis:
	Historical origins and modern communities of practice. CRC Press/Taylor and
	Francis, Boca Raton, FL.
	
	\bibitem[{Hoffman et~al.(1995)Hoffman, Shadbolt, Burton, and
		Klein}]{Hoffman1995}
	Hoffman, R.~R., Shadbolt, N.~R., Burton, A., Klein, G., 1995. Eliciting
	knowledge from experts: A methodological analysis. Organizational Behavior
	and Human Decision Processes 62~(2), 129 -- 158.
	
	\bibitem[{Holland(1986)}]{Holl1986}
	Holland, P.~W., 1986. Statistics and causal inference. Journal of the American
	Statistical Association 81~(396), 945--960.
	
	\bibitem[{Houghton et~al.(2015)Houghton, Baber, Stanton, Jenkins, and
		Revell}]{Houghton2015}
	Houghton, R.~J., Baber, C., Stanton, N.~A., Jenkins, D.~P., Revell, K., 2015.
	Combining network analysis with cognitive work analysis: insights into social
	organisational and cooperation analysis. Ergonomics, 1--16.
	
	\bibitem[{Huang et~al.(2011)Huang, Keisler, and Linkov}]{Huang2011}
	Huang, I.~B., Keisler, J., Linkov, I., 2011. Multi-criteria decision analysis
	in environmental sciences: Ten years of applications and trends. Science of
	The Total Environment 409~(19), 3578 -- 3594.
	
	\bibitem[{Ishibuchi and Nojima(2005)}]{Ishibuchi2005}
	Ishibuchi, H., Nojima, Y., May 2005. Comparison between fuzzy and interval
	partitions in evolutionary multiobjective design of rule-based classification
	systems. In: Fuzzy Systems, 2005. FUZZ '05. The 14th IEEE International
	Conference on. pp. 430--435.
	
	\bibitem[{Jagannathan and Elmaghraby(1985)}]{Jagannathan1985}
	Jagannathan, V., Elmaghraby, A.~S., 1985. Medkat: multiple expert delphi-based
	knowledge acquisition tool. In: Proceedings of the ACM NE Regional
	Conference. pp. 103--110.
	
	\bibitem[{Jagielska et~al.(1999)Jagielska, Matthews, and
		Whitfort}]{Jagielska1999}
	Jagielska, I., Matthews, C., Whitfort, T., 1999. An investigation into the
	application of neural networks, fuzzy logic, genetic algorithms, and rough
	sets to automated knowledge acquisition for classification problems.
	Neurocomputing 24~(1-3), 37 -- 54.
	
	\bibitem[{Jain et~al.(1999)Jain, Murty, and Flynn}]{Jain1999}
	Jain, A.~K., Murty, M.~N., Flynn, P.~J., Sep. 1999. Data clustering: A review.
	ACM Comput. Surv. 31~(3), 264--323.
	
	\bibitem[{Jansson et~al.(2015)Jansson, Erlandsson, and Axelsson}]{Jansson2015}
	Jansson, A., Erlandsson, M., Axelsson, A., 2015. Collegial verbalisation â€“
	the value of an independent observer: an ecological approach. Theoretical
	Issues in Ergonomics Science accepted - in press, 1--21.
	
	\bibitem[{Jonassen et~al.(2013)Jonassen, Beissner, and Yacci}]{Jonassen2013}
	Jonassen, D.~H., Beissner, K., Yacci, M., 2013. Structural knowledge:
	Techniques for representing, conveying, and acquiring structural knowledge.
	Routledge.
	
	\bibitem[{Jun et~al.(2012)Jun, Narayanan, Agarwal, Eddib, Singhal, Garimella,
		and Krovi}]{Jun2012}
	Jun, S.-k., Narayanan, M.~S., Agarwal, P., Eddib, A., Singhal, P., Garimella,
	S., Krovi, V., June 2012. Robotic minimally invasive surgical skill
	assessment based on automated video-analysis motion studies. In: Biomedical
	Robotics and Biomechatronics (BioRob), 2012 4th IEEE RAS EMBS International
	Conference on. pp. 25--31.
	
	\bibitem[{Kadhim et~al.(2014)Kadhim, Alam, and Kaur}]{Kadhim2014}
	Kadhim, M., Alam, M., Kaur, H., 2014. A multi-intelligent agent for knowledge
	discovery in database (miakdd): Cooperative approach with domain expert for
	rules extraction. In: Huang, D.-S., Jo, K.-H., Wang, L. (Eds.), Intelligent
	Computing Methodologies. Vol. 8589 of Lecture Notes in Computer Science.
	Springer International Publishing, pp. 602--614.
	
	\bibitem[{Kahn et~al.(1985)Kahn, Nowlan, and McDermott}]{Kahn1985}
	Kahn, G., Nowlan, S., McDermott, J., 1985. More: an intelligent knowledge
	acquisition tool. In: Proceedings of the Ninth International Conference on
	Artificial Intelligence. Los Angelos, California, pp. 581--584.
	
	\bibitem[{Kantardzic(2011)}]{Kantardzic2011}
	Kantardzic, M., 2011. Data mining: concepts, models, methods, and algorithms,
	2nd Edition. John Wiley \& Sons, NJ.
	
	\bibitem[{Kanungo et~al.(2002)Kanungo, Mount, Netanyahu, Piatko, Silverman, and
		Wu}]{Kanungo2002}
	Kanungo, T., Mount, D., Netanyahu, N., Piatko, C., Silverman, R., Wu, A., Jul
	2002. An efficient k-means clustering algorithm: analysis and implementation.
	Pattern Analysis and Machine Intelligence, IEEE Transactions on 24~(7),
	881--892.
	
	\bibitem[{Keeney(2013)}]{Keeney2013}
	Keeney, R.~L., 2013. Foundations for group decision analysis. Decision Analysis
	10~(2), 103--120.
	
	\bibitem[{Kieras(1988)}]{Kieras1988}
	Kieras, D.~E., 1988. Handbook of human-computer Interaction. Amsterdam:
	Elsevier, Ch. Towards a practical GOMS model methodology for user interface
	design.
	
	\bibitem[{Kim(2001)}]{Kim2001}
	Kim, I.~S., 2001. Human reliability analysis in the man-machine interface
	design review. Annals of Nuclear Energy 28~(11), 1069 -- 1081.
	
	\bibitem[{Kim et~al.(2010)Kim, McKay, and Moon}]{Kim2010}
	Kim, K., McKay, R.~B., Moon, B.-R., 2010. Multiobjective evolutionary
	algorithms for dynamic social network clustering. In: Proceedings of the 12th
	Annual Conference on Genetic and Evolutionary Computation. GECCO '10. ACM,
	New York, NY, USA, pp. 1179--1186.
	
	\bibitem[{Klein(2000)}]{Klein2000}
	Klein, G., 2000. Using cognitive task analysis to build a cognitive model.
	Proceedings of the Human Factors and Ergonomics Society Annual Meeting
	44~(6), 596--599.
	
	\bibitem[{Klein and Militello(1998)}]{Klein1998}
	Klein, G., Militello, L.~G., 1998. Cognitive task analysis. In: Workshop of
	human factors and ergonomics society 42nd Annual Meeting. Vol.~12. Chicago,
	Illinois.
	
	\bibitem[{Klein et~al.(1989)Klein, Calderwood, and MacGregor}]{Klein1989}
	Klein, G.~A., Calderwood, R., MacGregor, D., May 1989. Critical decision method
	for eliciting knowledge. Systems, Man and Cybernetics, IEEE Transactions on
	19~(3), 462--472.
	
	\bibitem[{Kober and Peters(2012)}]{Kober2012}
	Kober, J., Peters, J., 2012. Reinforcement learning in robotics: A survey. In:
	Wiering, M., van Otterlo, M. (Eds.), Reinforcement Learning. Vol.~12 of
	Adaptation, Learning, and Optimization. Springer Berlin Heidelberg, pp.
	579--610.
	
	\bibitem[{Kuehne(2013)}]{Kuehne2013}
	Kuehne, G., 2013. \"i don't know what's right anymore\": Engaging distressed
	interviewees using graphic-elicitation. Forum: Qualitative Social Research
	14~(3).
	
	\bibitem[{Kuo et~al.(2005)Kuo, Liao, and Tu}]{Kuo2005}
	Kuo, R., Liao, J., Tu, C., 2005. Integration of \{ART2\} neural network and
	genetic k-means algorithm for analyzing web browsing paths in electronic
	commerce. Decision Support Systems 40~(2), 355 -- 374.
	
	\bibitem[{Kurgan and Musilek(2006)}]{Kurgan2006}
	Kurgan, L.~A., Musilek, P., 2006. A survey of knowledge discovery and data
	mining process models. The Knowledge Engineering Review 21~(01), 1--24.
	
	\bibitem[{Kushniruk et~al.(2015)Kushniruk, Monkman, Tuden, Bellwood, and
		Borycki}]{Kushniruk2015}
	Kushniruk, A.~W., Monkman, H., Tuden, D., Bellwood, P., Borycki, E.~M., 2015.
	Integrating heuristic evaluation with cognitive walkthrough: development of a
	hybrid usability inspection method. Studies in health technology and
	informatics 208, 221--225.
	
	\bibitem[{Laird(2012)}]{Laird2012}
	Laird, J.~E., 2012. The soar cognitive architecture. AISB Quarterly 134, 1--4.
	
	\bibitem[{Lau et~al.(2008)Lau, Li, Song, and Kwok}]{Lau2008}
	Lau, R.~Y., Li, Y., Song, D., Kwok, R. C.~W., 2008. Knowledge discovery for
	adaptive negotiation agents in e-marketplaces. Decision Support Systems
	45~(2), 310 -- 323, i.T. and Value Creation.
	
	\bibitem[{Lee et~al.(2012)Lee, Seo, and Jung}]{Lee2012}
	Lee, D., Seo, H., Jung, M.~W., 2012. Neural basis of reinforcement learning and
	decision making. Annual review of neuroscience 35, 287--308.
	
	\bibitem[{Lemke(2012)}]{Lemke2012}
	Lemke, J., 2012. Analyzing Verbal Data: Principles, Methods, and Problems.
	Vol.~24 of Springer International Handbooks of Education. Springer
	Netherlands, book section~94, pp. 1471--1484.
	
	\bibitem[{Leu et~al.(2014{\natexlab{a}})Leu, Curtis, and Abbass}]{Leu2014a}
	Leu, G., Curtis, N.~J., Abbass, H.~A., 2014{\natexlab{a}}. Society of mind
	cognitive agent architecture applied to drivers adapting in a traffic
	context. Adaptive Behavior 22~(2), 123--145.
	
	\bibitem[{Leu et~al.(2014{\natexlab{b}})Leu, Tang, and Abbass}]{Leu2014}
	Leu, G., Tang, J., Abbass, H., 2014{\natexlab{b}}. On the role of working
	memory in trading-off skills and situation awareness in sudoku. In: Loo, C.,
	Yap, K., Wong, K., Beng~Jin, A., Huang, K. (Eds.), Neural Information
	Processing. Vol. 8836 of Lecture Notes in Computer Science. Springer
	International Publishing, pp. 571--578.
	
	\bibitem[{Li et~al.(2012)Li, Kannry, Kushniruk, Chrimes, McGinn, Edonyabo, and
		Mann}]{Li2012}
	Li, A.~C., Kannry, J.~L., Kushniruk, A., Chrimes, D., McGinn, T.~G., Edonyabo,
	D., Mann, D.~M., 2012. Integrating usability testing and think-aloud protocol
	analysis with near-live clinical simulations in evaluating clinical decision
	support. International Journal of Medical Informatics 81~(11), 761 -- 772.
	
	\bibitem[{Linster(1993)}]{Linster1993}
	Linster, M., 1993. Explicit and operational models as a basis for second
	generation knowledge acquisition tools. In: David, J.-M., Krivine, J.-P.,
	Simmons, R. (Eds.), Second Generation Expert Systems. Springer Berlin
	Heidelberg, pp. 465--494.
	
	\bibitem[{Liu et~al.(2006)Liu, Feyen, and Tsimhoni}]{Liu2006}
	Liu, Y., Feyen, R., Tsimhoni, O., Mar. 2006. Queueing network-model human
	processor (qn-mhp): A computational architecture for multitask performance in
	human-machine systems. ACM Trans. Comput.-Hum. Interact. 13~(1), 37--70.
	
	\bibitem[{Lotem and Halpern(2012)}]{Lotem2012}
	Lotem, A., Halpern, J.~Y., 2012. Coevolution of learning and data-acquisition
	mechanisms: a model for cognitive evolution. Philosophical Transactions of
	the Royal Society of London B: Biological Sciences 367~(1603), 2686--2694.
	
	\bibitem[{Mahatody et~al.(2010)Mahatody, Sagar, and Kolski}]{Mahatody2010}
	Mahatody, T., Sagar, M., Kolski, C., 2010. State of the art on the cognitive
	walkthrough method, its variants and evolutions. International Journal of
	Human-Computer Interaction 2~(8), 741--785.
	
	\bibitem[{Marcus and McDermott(1989)}]{Marcus1989}
	Marcus, S., McDermott, J., 1989. Salt: A knowledge acquisition language for
	propose-and-revise systems. Artificial Intelligence 39~(1), 1 -- 37.
	
	\bibitem[{Marsland(2014)}]{Marsland2014}
	Marsland, S., 2014. Machine learning: an algorithmic perspective, 2nd Edition.
	Macine Learning \& Pattern Recognition. CRC press, FL.
	
	\bibitem[{McCloskey et~al.(1991)McCloskey, Geiwitz, and
		Kornell}]{McCloskey1991}
	McCloskey, B.~P., Geiwitz, J., Kornell, J., 1991. Empirical comparisons of
	knowledge acquisition techniques. Proceedings of the Human Factors and
	Ergonomics Society Annual Meeting 35~(5), 268--272.
	
	\bibitem[{McNelis et~al.(2014)McNelis, Ironside, Zvonar, and
		Ebright}]{McNelis2014}
	McNelis, A., Ironside, P., Zvonar, S., Ebright, P., 2014. Advancing the science
	of research in nursing education: Contributions of the critical decision
	method. J Nurs Educ 53~(2), 61--64.
	
	\bibitem[{Meyer(1992)}]{Meyer1992}
	Meyer, M.~A., Sep 1992. How to apply the anthropological technique of
	participant observation to knowledge acquisition for expert systems. Systems,
	Man and Cybernetics, IEEE Transactions on 22~(5), 983--991.
	
	\bibitem[{Militello and Hoffman(2008)}]{Militello2008}
	Militello, L.~G., Hoffman, R.~R., 2008. The forgotten history of cognitive task
	analysis. Proceedings of the Human Factors and Ergonomics Society Annual
	Meeting 52~(4), 383--387.
	
	\bibitem[{Militello and Hutton(1998)}]{Militello1998}
	Militello, L.~G., Hutton, R. J.~B., 1998. Applied cognitive task analysis
	(acta): a practitioner's toolkit for understanding cognitive task demands.
	Ergonomics 41~(11), 1618--1641.
	
	\bibitem[{Mohri et~al.(2012)Mohri, Rostamizadeh, and Talwalkar}]{Mohri2012}
	Mohri, M., Rostamizadeh, A., Talwalkar, A., 2012. Foundations of machine
	learning. MIT press.
	
	\bibitem[{Moradi et~al.(2013)Moradi, Aghaie, and Hosseini}]{Moradi2013}
	Moradi, M., Aghaie, A., Hosseini, M., 2013. Knowledge-collector agents:
	Applying intelligent agents in marketing decisions with knowledge management
	approach. Knowledge-Based Systems 52~(0), 181 -- 193.
	
	\bibitem[{Mukhopadhyay and Maulik(2011)}]{Mukhopadhyay2011}
	Mukhopadhyay, A., Maulik, U., 2011. A multiobjective approach to \{MR\} brain
	image segmentation. Applied Soft Computing 11~(1), 872 -- 880.
	
	\bibitem[{Mukhopadhyay et~al.(2014{\natexlab{a}})Mukhopadhyay, Maulik,
		Bandyopadhyay, and Coello}]{Mukhopadhyay2014a}
	Mukhopadhyay, A., Maulik, U., Bandyopadhyay, S., Coello, C., Feb
	2014{\natexlab{a}}. Survey of multiobjective evolutionary algorithms for data
	mining: Part ii. Evolutionary Computation, IEEE Transactions on 18~(1),
	20--35.
	
	\bibitem[{Mukhopadhyay et~al.(2014{\natexlab{b}})Mukhopadhyay, Maulik,
		Bandyopadhyay, and Coello~Coello}]{Mukhopadhyay2014}
	Mukhopadhyay, A., Maulik, U., Bandyopadhyay, S., Coello~Coello, C., Feb
	2014{\natexlab{b}}. A survey of multiobjective evolutionary algorithms for
	data mining: Part i. Evolutionary Computation, IEEE Transactions on 18~(1),
	4--19.
	
	\bibitem[{Naweed(2014)}]{Naweed2014}
	Naweed, A., 2014. Investigations into the skills of modern and traditional
	train driving. Applied Ergonomics 45~(3), 462 -- 470.
	
	\bibitem[{Neale(1988)}]{Neale1988}
	Neale, I.~M., 6 1988. First generation expert systems: a review of knowledge
	acquisition methodologies. The Knowledge Engineering Review 3, 105--145.
	
	\bibitem[{Neapolitan(2012)}]{Neapolitan2012}
	Neapolitan, R.~E., 2012. Probabilistic Reasoning In Expert Systems: Theory and
	Algorithms. CreateSpace Independent Publishing Platform, USA.
	
	\bibitem[{Ngai et~al.(2009)Ngai, Xiu, and Chau}]{Ngai2009}
	Ngai, E. W.~T., Xiu, L., Chau, D. C.~K., 2009. Application of data mining
	techniques in customer relationship management: A literature review and
	classification. Expert Systems with Applications 36~(2, Part 2), 2592 --
	2602.
	
	\bibitem[{Nguyen et~al.(2013)Nguyen, Luu, Poch, and Thompson}]{Nguyen2013}
	Nguyen, H., Luu, T.-D., Poch, O., Thompson, J.~D., 2013. Knowledge discovery in
	variant databases using inductive logic programming. Bioinformatics and
	Biology Insights 7, 119--131.
	
	\bibitem[{Notermans and Kommers(2013)}]{Notermans2013}
	Notermans, C., Kommers, H., 2013. Researching religion: the iconographic
	elicitation method. Qualitative Research 13~(5), 608--625.
	
	\bibitem[{Okoli et~al.(2013)Okoli, Weller, Watt, and Wong}]{Okoli2013}
	Okoli, J.~O., Weller, G., Watt, J., Wong, B. L.~W., 2013. Decision making
	strategies used by experts and the potential for training intuitive skills: a
	preliminary study. In: Chaudet, H., Pellegrin, L., Bonnardel, N. (Eds.), The
	11th International Conference on Naturalistic Decision Making - NDM 2013.
	Arpege Science Publishing.
	
	\bibitem[{Olson and Biolsi(1991)}]{Olson1991}
	Olson, J.~R., Biolsi, K.~J., 1991. Study of Expertise: Prospects and Limits.
	Cambridge University Press, Ch. Techniques for representing knowledge
	structures, pp. 240--285.
	
	\bibitem[{Orasanu(2001)}]{Orasanu2001}
	Orasanu, J., 2001. Decision making (naturalistic), psychology of. In: Baltes,
	N. J. S.~B. (Ed.), International Encyclopedia of the Social \& Behavioral
	Sciences. Pergamon, Oxford, pp. 3300 -- 3304.
	
	\bibitem[{Orru et~al.(2012)Orru, Pettersson-Yeo, Marquand, Sartori, and
		Mechelli}]{Orru2012}
	Orru, G., Pettersson-Yeo, W., Marquand, A.~F., Sartori, G., Mechelli, A., 2012.
	Using support vector machine to identify imaging biomarkers of neurological
	and psychiatric disease: A critical review. Neuroscience \& Biobehavioral
	Reviews 36~(4), 1140 -- 1152.
	
	\bibitem[{Overbey et~al.(2010)Overbey, McKoy, Gordon, and
		McKitrick}]{Overbey2010}
	Overbey, L.~A., McKoy, G., Gordon, J., McKitrick, S., May 2010. Automated
	sensing and social network analysis in virtual worlds. In: Intelligence and
	Security Informatics (ISI), 2010 IEEE International Conference on. pp.
	179--184.
	
	\bibitem[{Park and Park(2012)}]{Park2012}
	Park, K., Park, S., 2012. Development of professional engineers' authentic
	contexts in blended learning environments. British Journal of Educational
	Technology 43~(1), E14--E18.
	
	\bibitem[{Pauley et~al.(2013)Pauley, Flin, and Azuara-Blanco}]{Pauley2013}
	Pauley, K., Flin, R., Azuara-Blanco, A., 2013. Intra-operative decision making
	by ophthalmic surgeons. British Journal of Ophthalmology.
	
	\bibitem[{Pauley et~al.(2011)Pauley, Flin, Yule, and Youngson}]{Pauley2011}
	Pauley, K., Flin, R., Yule, S., Youngson, G., 2011. Surgeons' intraoperative
	decision making and risk management. The American Journal of Surgery 202~(4),
	375 -- 381.
	
	\bibitem[{Payne and Green(1989)}]{Payne1989}
	Payne, S.~J., Green, T., 1989. The structure of command languages: an
	experiment on task-action grammar. International Journal of Man-Machine
	Studies 30~(2), 213 -- 234.
	
	\bibitem[{Perry and Krippendorff(2013)}]{Perry2013}
	Perry, G.~T., Krippendorff, K., 2013. On the reliability of identifying design
	moves in protocol analysis. Design Studies 34~(5), 612 -- 635.
	
	\bibitem[{Petraki and Abbass(2014)}]{Petraki2014}
	Petraki, E., Abbass, H., Dec 2014. On trust and influence: A computational red
	teaming game theoretic perspective. In: Computational Intelligence for
	Security and Defense Applications (CISDA), 2014 Seventh IEEE Symposium on.
	pp. 1--7.
	
	\bibitem[{Petrovic and Brand(2009)}]{Petrovic2009}
	Petrovic, O., Brand, A., 2009. Serious games on the move. Springer.
	
	\bibitem[{Pidgeon et~al.(1991)Pidgeon, Turner, and Blockley}]{Pidgeon1991}
	Pidgeon, N., Turner, B.~A., Blockley, D., 1991. The use of grounded theory for
	conceptual analysis in knowledge elicitation. International Journal of
	Man-Machine Studies 35, 151--173.
	
	\bibitem[{Plant and Stanton(2013)}]{Plant2013}
	Plant, K.~L., Stanton, N.~A., 2013. What is on your mind? using the perceptual
	cycle model and critical decision method to understand the decision-making
	process in the cockpit. Ergonomics 56~(8), 1232--1250, pMID: 23800131.
	
	\bibitem[{Plant and Stanton(2014)}]{Plant2014}
	Plant, K.~L., Stanton, N.~A., 2014. The process of processing: exploring the
	validity of neisser's perceptual cycle model with accounts from critical
	decision-making in the cockpit. Ergonomics 0~(accepted-in press), 1--15.
	
	\bibitem[{Powell et~al.(2015)Powell, Guadagno, and Benson}]{Powell2015}
	Powell, M.~B., Guadagno, B., Benson, M., 2015. Improving child investigative
	interviewer performance through computer-based learning activities. Policing
	and Society accepted - in press~(0), 1--10.
	
	\bibitem[{Pu et~al.(2011)Pu, Faltings, Chen, Zhang, and Viappiani}]{Pu2011}
	Pu, P., Faltings, B., Chen, L., Zhang, J., Viappiani, P., 2011. Usability
	Guidelines for Product Recommenders Based on Example Critiquing Research.
	Springer US, book section~16, pp. 511--545.
	
	\bibitem[{Pugh et~al.(2011)Pugh, DaRosa, Santacaterina, and Clark}]{Pugh2011}
	Pugh, C., DaRosa, D., Santacaterina, S., Clark, R., 2011. Faculty evaluation of
	simulation-based modules for assessment of intraoperative decision making.
	Surgery 149~(4), 534 -- 542.
	
	\bibitem[{Ragsdell et~al.(2014)Ragsdell, Probets, Ahmed, and
		Murray}]{Ragsdell2014}
	Ragsdell, G., Probets, S., Ahmed, G., Murray, I., 2014. Knowledge audit:
	Findings from the energy sector. Knowledge and Process Management 21~(4),
	270--279.
	
	\bibitem[{Rahman and Shukor(2011)}]{Rahman2011}
	Rahman, A., Shukor, N., Nov 2011. Knowledge audit process - tales of two
	organizations. In: Research and Innovation in Information Systems (ICRIIS),
	2011 International Conference on. pp. 1--5.
	
	\bibitem[{Ralha and Silva(2012)}]{Ralha2012}
	Ralha, C.~G., Silva, C. V.~S., 2012. A multi-agent data mining system for
	cartel detection in brazilian government procurement. Expert Systems with
	Applications 39~(14), 11642 -- 11656.
	
	\bibitem[{Rebentrost et~al.(2014)Rebentrost, Mohseni, and
		Lloyd}]{Rebentrost2014}
	Rebentrost, P., Mohseni, M., Lloyd, S., Sep 2014. Quantum support vector
	machine for big data classification. Phys. Rev. Lett. 113, 130503.
	
	\bibitem[{Reed and Rowe(2004)}]{Reed2004}
	Reed, C., Rowe, G., 2004. Araucaria: software for argument analysis,
	diagramming and representation. International Journal on Artificial
	Intelligence Tools 13~(04), 961--979.
	
	\bibitem[{Reisner(1981)}]{Reisner1981}
	Reisner, P., March 1981. Formal grammar and human factors design of an
	interactive graphics system. Software Engineering, IEEE Transactions on
	SE-7~(2), 229--240.
	
	\bibitem[{Richard and Lahman(2015)}]{Richard2015}
	Richard, V.~M., Lahman, M. K.~E., 2015. Photo-elicitation: reflexivity on
	method, analysis, and graphic portraits. International Journal of Research \&
	Method in Education 38~(1), 3--22.
	
	\bibitem[{Rokach(2007)}]{Rokach2007}
	Rokach, L., 2007. Data mining with decision trees: theory and applications.
	Machine Perception and Artificial Intelligence. World scientific.
	
	\bibitem[{Roth(2008)}]{Roth2008}
	Roth, E.~M., 2008. Uncovering the requirements of cognitive work. Human
	Factors: The Journal of the Human Factors and Ergonomics Society 50~(3),
	475--480.
	
	\bibitem[{Roth et~al.(2014)Roth, O'Hara, Bisantz, Endsley, Hoffman, Klein,
		Militello, and Pfautz}]{Roth2014}
	Roth, E.~M., O'Hara, J., Bisantz, A., Endsley, M.~R., Hoffman, R., Klein, G.,
	Militello, L., Pfautz, J.~D., 2014. Discussion panel: How to recognize a
	``good'' cognitive task analysis? Proceedings of the Human Factors and
	Ergonomics Society Annual Meeting 58~(1), 320--324.
	
	\bibitem[{Roth et~al.(1992)Roth, Woods, and Pople}]{Roth1992}
	Roth, E.~M., Woods, D.~D., Pople, H.~E., 1992. Cognitive simulation as a tool
	for cognitive task analysis. Ergonomics 35~(10), 1163--1198.
	
	\bibitem[{Rugg et~al.(1992)Rugg, Corbridge, Major, Burton, and
		Shadbolt}]{Rugg1992}
	Rugg, G., Corbridge, C., Major, N., Burton, A., Shadbolt, N., 1992. A
	comparison of sorting techniques in knowledge acquisition. Knowledge
	Acquisition 4~(3), 279--291.
	
	\bibitem[{Rybakov(2009)}]{Rybakov2009}
	Rybakov, V., 2009. Logic of knowledge and discovery via interacting agents -
	decision algorithm for true and satisfiable statements. Information Sciences
	179~(11), 1608 -- 1614, including Special Issue on Chance Discovery Discovery
	of Significant Events for Decision.
	
	\bibitem[{Ryder and Redding(1993)}]{Ryder1993}
	Ryder, J.~M., Redding, R.~E., 1993. Integrating cognitive task analysis into
	instructional systems development. Educational Technology Research and
	Development 41~(2), 75--96.
	
	\bibitem[{Ryder and Zachary(1991)}]{Ryder1991}
	Ryder, J.~M., Zachary, W.~W., 1991. Experimental validation of the attention
	switching component of the cognet framework. Proceedings of the Human Factors
	and Ergonomics Society Annual Meeting 35~(2), 72--76.
	
	\bibitem[{Salmon et~al.(2012)Salmon, Pipe, Malloy, and Mackay}]{Salmon2012}
	Salmon, K., Pipe, M.-E., Malloy, A., Mackay, K., 2012. Do non-verbal aids
	increase the effectiveness of â€˜best practiceâ€™ verbal interview
	techniques? an experimental study. Applied Cognitive Psychology 26~(3),
	370--380.
	
	\bibitem[{Salvucci(2013)}]{Salvucci2013}
	Salvucci, D.~D., 2013. Integration and reuse in cognitive skill acquisition.
	Cognitive Science 37~(5), 829--860.
	
	\bibitem[{Schraagen et~al.(2000)Schraagen, Chipman, and Shalin}]{Schraagen2000}
	Schraagen, J.~M., Chipman, S.~F., Shalin, V.~L., 2000. Cognitive task analysis.
	Mahwah, NJ: Lawrence Erlbaum Associates.
	
	\bibitem[{Schraagen et~al.(2008)Schraagen, Ormerod, Militello, and
		Lipshitz}]{Schraagen2008}
	Schraagen, J.~M., Ormerod, T., Militello, L., Lipshitz, R., 2008. Naturalistic
	decision making and macrocognition. Ashgate Publishing, Ltd., Aldershot, UK.
	
	\bibitem[{Schroeder and Bazzan(2002)}]{Schroeder2002}
	Schroeder, L.~F., Bazzan, A. L.~C., 2002. A multi-agent system to facilitate
	knowledge discovery: an application to bioinformatics. In: Workshop on
	Bioinformatics and Multi-Agent Systems. pp. 44--50.
	
	\bibitem[{Seager et~al.(2011)Seager, Ruskov, Sasse, and Oliveira}]{Seager2011}
	Seager, W., Ruskov, M., Sasse, M.~A., Oliveira, M., 2011. Eliciting and
	modelling expertise for serious games in project management. Entertainment
	Computing 2~(2), 75 -- 80.
	
	\bibitem[{Secretan et~al.(2010)Secretan, Georgiopoulos, Koufakou, and
		Cardona}]{Secretan2010}
	Secretan, J., Georgiopoulos, M., Koufakou, A., Cardona, K., 2010. Aphid: An
	architecture for private, high-performance integrated data mining. Future
	Generation Computer Systems 26~(7), 891 -- 904.
	
	\bibitem[{Sharples et~al.(2002)Sharples, Jeffery, Du~Boulay, Teather, Teather,
		and Du~Boulay}]{Sharples2002}
	Sharples, M., Jeffery, N., Du~Boulay, J. B.~H., Teather, D., Teather, B.,
	Du~Boulay, G.~H., 2002. Socio-cognitive engineering: a methodology for the
	design of human-centred technology. European Journal of Operational Research
	136~(2), 310--323.
	
	\bibitem[{Sowa(2014)}]{Sowa2014}
	Sowa, J.~F., 2014. Conceptual analysis as a basis for knowledge acquisition.
	In: Hoffman, R.~R. (Ed.), The Psychology of Expertise. Psychology Press New
	York, pp. 80--96.
	
	\bibitem[{Stanton et~al.(2013)Stanton, M., Walker, Rafferty, Baber, and
		Jenkins}]{Stanton2013}
	Stanton, N.~A., M., S.~P., Walker, G.~H., Rafferty, L.~A., Baber, C., Jenkins,
	D.~P., 2013. Human factors methods: a practical guide for engineering and
	design, 2nd Edition. Ashgate Publishing, Ltd.
	
	\bibitem[{Steidtmann et~al.(2013)Steidtmann, Manber, Blasey, Markowitz, Klein,
		Rothbaum, Thase, Kocsis, and Arnow}]{Steidtmann2013}
	Steidtmann, D., Manber, R., Blasey, C., Markowitz, J.~C., Klein, D.~N.,
	Rothbaum, B.~O., Thase, M.~E., Kocsis, J.~H., Arnow, B.~A., 2013. Detecting
	critical decision points in psychotherapy and psychotherapy + medication for
	chronic depression. Journal of Consulting and Clinical Psychology 81~(5),
	783--792.
	
	\bibitem[{Sun(2006)}]{Sun2006}
	Sun, R., 2006. The CLARION cognitive architecture: Extending cognitive modeling
	to social simulation. Cambridge University Press, New York.
	
	\bibitem[{Taylor(1967)}]{Taylor1967}
	Taylor, F.~W., 1967. The principles of scientific management. Norton, New York.
	
	\bibitem[{Tofel-Grehl and Feldon(2013)}]{Tofel-Grehl2013}
	Tofel-Grehl, C., Feldon, D.~F., 2013. Cognitive task analysis-based training: A
	meta-analysis of studies. Journal of Cognitive Engineering and Decision
	Making 7~(3), 293--304.
	
	\bibitem[{Varga-Atkins and O'Brien(2009)}]{Varga-Atkins2009}
	Varga-Atkins, T., O'Brien, M., 2009. From drawings to diagrams: maintaining
	researcher control during graphic elicitation in qualitative interviews.
	International Journal of Research \& Method in Education 32~(1), 53--67.
	
	\bibitem[{Vatolkin(2015)}]{Vatolkin2015}
	Vatolkin, I., 2015. Exploration of two-objective scenarios on supervised
	evolutionary feature selection: A survey and a case study (application to
	music categorisation). In: Gaspar-Cunha, A., Henggeler~Antunes, C., Coello,
	C.~C. (Eds.), Evolutionary Multi-Criterion Optimization. Vol. 9019 of Lecture
	Notes in Computer Science. Springer International Publishing, pp. 529--543.
	
	\bibitem[{Wang et~al.(2013)Wang, Shafi, Lokan, and Abbass}]{Wang2013}
	Wang, S.~L., Shafi, K., Lokan, C., Abbass, H.~A., 2013. An agent-based model to
	simulate and analyse behaviour under noisy and deceptive information.
	Adaptive Behavior 21~(2), 96--117.
	
	\bibitem[{Wannheden et~al.(2013)Wannheden, Westling, Savage, Sandahl, and
		Ellenius}]{Wannheden2013}
	Wannheden, C., Westling, K., Savage, C., Sandahl, C., Ellenius, J., 2013. Hiv
	and tuberculosis coinfection: a qualitative study of treatment challenges
	faced by care providers. The International Journal of Tuberculosis and Lung
	Disease 17~(8), 1029--1035.
	
	\bibitem[{Ward et~al.(2011)Ward, Suss, Eccles, Williams, and Harris}]{Ward2011}
	Ward, P., Suss, J., Eccles, D., Williams, A., Harris, K., 2011. Skill-based
	differences in option generation in a complex task: a verbal protocol
	analysis. Cognitive Processing 12~(3), 289--300.
	
	\bibitem[{Wei and Salvendy(2004)}]{Wei2004}
	Wei, J., Salvendy, G., 2004. The cognitive task analysis methods for job and
	task design: review and reappraisal. Behaviour \& Information Technology
	23~(4), 273--299.
	
	\bibitem[{Woodward(1990)}]{Woodward1990}
	Woodward, B., 1990. Knowledge acquisition at the front end: defining the
	domain. Knowledge Acquisition 2~(1), 73 -- 94.
	
	\bibitem[{Yagahara et~al.(2013)Yagahara, Tsuji, Fukuda, Yokooka, Nishimoto,
		Kurowarabi, and Ogasawara}]{Yagahara2013}
	Yagahara, A., Tsuji, S., Fukuda, A., Yokooka, Y., Nishimoto, N., Kurowarabi,
	K., Ogasawara, K., 2013. Constructing mammography examination process
	ontology using affinity diagram and hierarchical task analysis. Studies in
	health technology and informatics 192, 1059.
	
	\bibitem[{Yates(2007)}]{Yates2007}
	Yates, K., 2007. Towards a taxonomy of cognitive task analysis moethods: a
	search for cognition and task analysis interactions. Thesis.
	
	\bibitem[{Yates and Feldon(2011)}]{Yates2011}
	Yates, K.~A., Feldon, D.~F., 2011. Advancing the practice of cognitive task
	analysis: a call for taxonomic research. Theoretical Issues in Ergonomics
	Science 12~(6), 472--495.
	
	\bibitem[{Yee and Bailenson(2008)}]{Yee2008}
	Yee, N., Bailenson, J.~N., 2008. A method for longitudinal behavioral data
	collection in second life. Presence: Teleoperators and Virtual Environments
	17~(6), 594--596.
	
	\bibitem[{Yoo et~al.(2013)Yoo, Gretzel, and Zanker}]{Yoo2013}
	Yoo, K.-H., Gretzel, U., Zanker, M., 2013. Implications for Recommender System
	Design. SpringerBriefs in Electrical and Computer Engineering. Springer New
	York, book section~7, pp. 37--44.
	
	\bibitem[{Yusoff and Salim(2012)}]{Yusoff2012}
	Yusoff, N.~M., Salim, S.~S., 2012. Investigating cognitive task difficulties
	and expert skills in e-learning storyboards using a cognitive task analysis
	technique. Computers \& Education 58~(1), 652 -- 665.
	
	\bibitem[{Zhou et~al.(2011)Zhou, Qu, Li, Zhao, Suganthan, and Zhang}]{Zhou2011}
	Zhou, A., Qu, B.-Y., Li, H., Zhao, S.-Z., Suganthan, P.~N., Zhang, Q., 2011.
	Multiobjective evolutionary algorithms: A survey of the state of the art.
	Swarm and Evolutionary Computation 1~(1), 32 -- 49.
	
	\bibitem[{Zhou et~al.(2010)Zhou, Rao, and Lv}]{Zhou2010}
	Zhou, D., Rao, W., Lv, F., Dec 2010. A multi-agent distributed data mining
	model based on algorithm analysis and task prediction. In: 2nd International
	Conference on Information Engineering and Computer Science. pp. 1--4.
	
\end{thebibliography}

\end{document}